\documentclass[]{commtr}

\usepackage{bm}
\usepackage{algorithm}
\usepackage{algpseudocode}
\usepackage{setspace}

\usepackage{color,soul}

\commtrsetup{
title    = {COIN: Collaborative Interaction-Aware Multi-Agent Reinforcement Learning for Self-Driving Systems},
author   = {Yifeng Zhang$^{\rm a}$, Jieming Chen$^{\rm b}$, Tingguang Zhou$^{\rm a}$, Tanishq Duhan$^{\rm a}$, Jianghong Dong$^{\rm c, *}$, Yuhong Cao$^{\rm a}$, Guillaume Sartoretti$^{\rm a}$},
address  = {
\tdd{a} Department of Mechanical Engineering, College of Design and Engineering,  National University of Singapore, 21 Lower Kent Ridge Rd, 119077, Singapore \sep
\tdd{b} Department of Electrical and Electronic Engineering, The Hong Kong Polytechnic University, Hong Kong, 100872, China \sep
\tdd{c} School of Vehicle and Mobility, Tsinghua University, Beijing, 100084, China \sep
\tdd{*} Corresponding author.
},
email    = {yifeng@u.nus.edu, dong-jianghong@nus.edu.sg}, 
abstract = {
\textbf{
Multi-Agent Self-Driving (MASD) systems provide an effective solution for coordinating autonomous vehicles to reduce congestion and enhance both safety and operational efficiency in future intelligent transportation systems. 
Multi-Agent Reinforcement Learning (MARL) has emerged as a promising approach for developing advanced end-to-end MASD systems.  
However, achieving efficient and safe collaboration in dynamic MASD systems remains a significant challenge in dense scenarios with complex agent interactions.
To address this challenge, we propose a novel collaborative(CO-) interaction-aware(-IN) MARL framework, named COIN. 
Specifically, we develop a new counterfactual individual-global twin delayed deep deterministic policy gradient (CIG-TD3) algorithm, crafted in a ``centralized training, decentralized execution'' (CTDE) manner, which aims to jointly optimize the individual objectives (navigation) and the global objectives (collaboration) of agents. 
We further introduce a dual-level interaction-aware centralized critic architecture that captures both local pairwise interactions and global system-level dependencies, enabling more accurate global value estimation and improved credit assignment for collaborative policy learning.
We conduct extensive simulation experiments in dense urban traffic environments, which demonstrate that COIN consistently outperforms other advanced baseline methods in both safety and efficiency across various system sizes. 
These results highlight its superiority in complex and dynamic MASD scenarios, as further validated through real-world robot demonstrations. 
Supplementary videos are available at \href{https://marmotlab.github.io/COIN/}{https://marmotlab.github.io/COIN/}.
}
},
keywords = {
Multi-agent reinforcement learning \sep Multi-agent self-driving system \sep End-to-end navigation
},
}

\begin{document}

\maketitle  

\section{Introduction}
With the rapid development of artificial intelligence and sensing technologies, autonomous vehicles (AVs) have attracted worldwide attention and are becoming an important part of future intelligent transportation systems (ITS)~\cite{haydari2020deep}.
They have shown great promise in improving transportation systems by enhancing driving safety, reducing congestion, and increasing fuel efficiency, enabled by advanced sensors and intelligent decision-making algorithms~\cite{chib2023recent, chen2024end}.
Unlike traditional modular autonomous driving frameworks, which separate driving tasks into perception, planning, and control, end-to-end frameworks learn to directly map raw sensor inputs to driving decisions in a single integrated pipeline~\cite{tampuu2020survey, hu2023planning, huang2023differentiable}.
This design provides higher flexibility and adaptability in complex, dynamic, and diverse driving scenarios, offering a more unified and robust solution for building efficient autonomous driving systems.

The navigation problem in end-to-end self-driving systems composed solely of AVs involves determining the optimal path and corresponding low-level control commands to reach a destination safely and efficiently, traditionally solved through rule-based algorithms and pre-defined models~\cite{treiber2000congested, kesting2007general, li2018consensus, chen2022cooperation, cai2022formation}. 
However, these methods often rely on explicit modeling assumptions, such as simplified vehicle dynamics or decoupled control designs, which are difficult to maintain in dynamic traffic scenarios with dense and strongly coupled interactions.
Data-driven methods, such as Reinforcement Learning (RL), provide a promising alternative by learning fast-responsive control strategies through interaction with the environment, without relying on explicit models of system dynamics.
This enables AVs to adapt to dynamic urban conditions, and handle complex scenarios that are difficult to model with traditional approaches.

Single-agent RL provides a simple formulation by learning a single policy under the assumption that the environment remains stationary (i.e., its transition dynamics depend only on the agent’s own actions).
However, such methods often struggle in dynamic Multi-Agent Self-Driving (MASD) systems (i.e., multi-vehicle scenarios) due to the complexity of interactions and coordination among vehicles.
Multi-Agent Reinforcement Learning (MARL)~\cite{yang2023learning, han2020cooperative, rashid2020monotonic, foerster2018counterfactual} addresses these challenges by enabling agents to learn cooperative behaviors in a decentralized manner, allowing them to coordinate efficiently while naturally scaling to larger numbers of agents.
Although MARL introduces additional challenges such as non-stationary learning dynamics and multi-agent credit assignment, its ability to model interdependent decision-making makes it a more suitable paradigm for MASD systems.
Recent MARL work has moved beyond discrete action spaces and explored collaborative learning in low-level continuous control domains, which enables more fine-grained and adaptive driving behaviors, thereby improving overall safety and efficiency.
For example, CoPO~\cite{peng2021learning} introduced a bi-level optimization framework that learns a coordination factor to weight ego and neighbor rewards during policy updates, with this factor optimized via meta-gradients from the global reward, which serves sorely as a meta-level signal rather than a directly optimized objective.
More recently, SAPO~\cite{dai2023socially} proposed an attention-based social-aware module to dynamically identify the most relevant neighbor and align their objectives, but its optimization remained limited to pairwise interactions.
TraCo~\cite{liu2023traco} employed cross-attention to model dynamic inter-agent relations and used a counterfactual advantage to facilitate coordination among neighboring agents, yet its optimization remained limited to the local level without addressing global coordination.
While these methods have advanced local coordination, interaction modeling, and socially-aware behavior, they still fall short in capturing complex multi-agent interactions and in explicitly optimizing for global objectives.

\begin{figure}[t!]
    \centering
    \includegraphics[width=\linewidth]{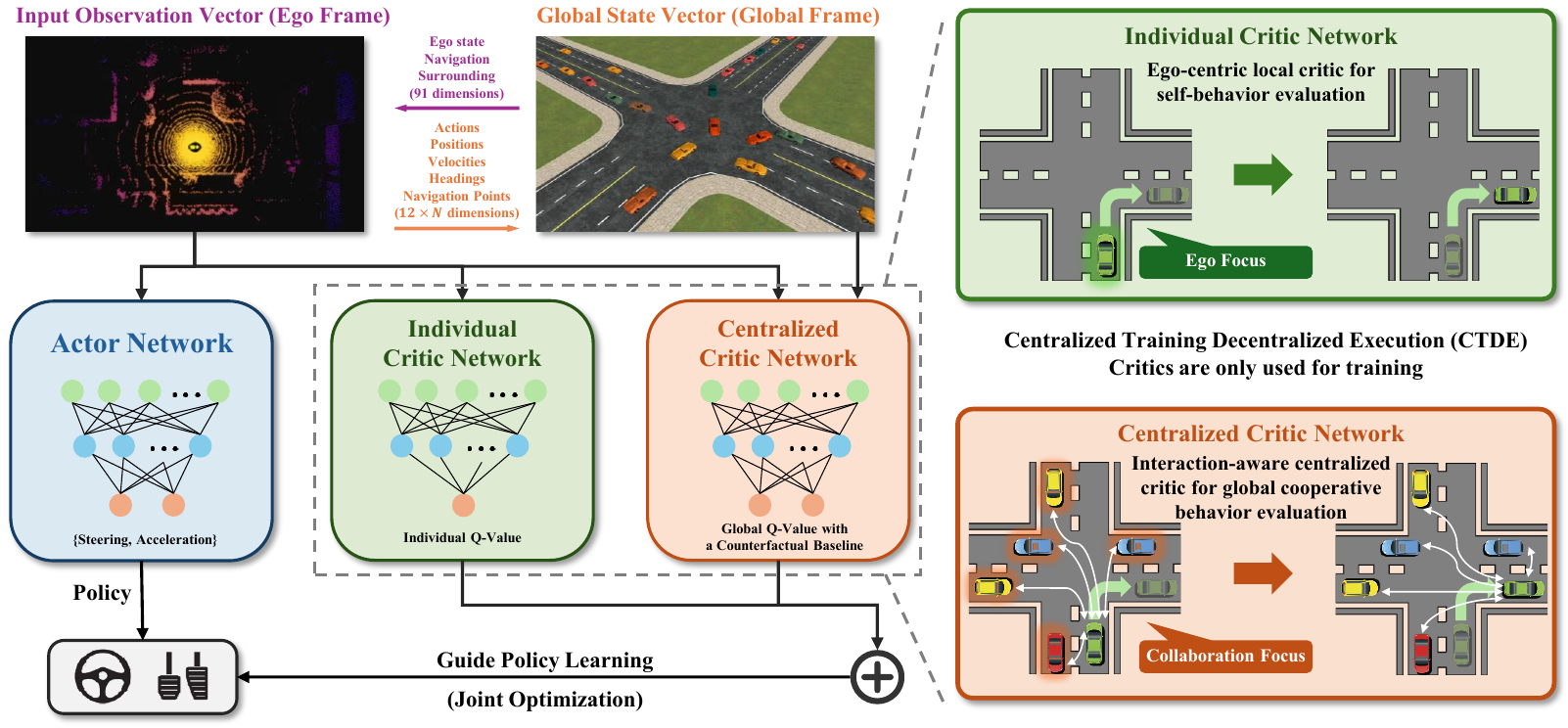}
    \caption{The overall COIN framework, which follows the CTDE paradigm. During training, an individual critic learns individual Q-values to guide the ego vehicle’s navigation policy, while a centralized critic learns global Q-values to facilitate collaborative policy learning. These two critics work together to achieve joint policy optimization. During execution, the actor learns an end-to-end policy based only on local observations. The right side illustrates the roles of the individual critic and centralized critic. Notably, we design a dual-level interaction-aware centralized critic with a counterfactual baseline to improve credit assignment and promote better agent cooperation.}
    \label{fig:intro}
\end{figure}

To address these challenges, we propose a novel Collaborative (CO-) and Interaction-aware (-IN) MARL framework, named COIN, designed for end-to-end navigation for MASD systems under dense and highly interactive traffic environments.
Our overall COIN framework is illustrated in Fig.~\ref{fig:intro}.
Particularly, we introduce a novel counterfactual individual-global twin delayed deep deterministic policy gradient (CIG-TD3) algorithm under the centralized training and decentralized execution (CTDE) paradigm, which jointly optimizes both local (navigation) and global (collaboration) objectives.
In contrast to individual learning algorithms, which only use a local critic to guide policy learning, CIG-TD3 incorporates an additional centralized critic to estimate global state-action values, thus enabling the joint optimization of individual and global objectives.
Moreover, it jointly optimizes local and global critics in parallel, rather than decomposing the global value into local terms, allowing them to play complementary roles guiding ego-agent navigation and multi-agent collaboration.
Unlike previous MARL methods for MASD systems~\cite{peng2021learning, liu2023traco, dai2023socially}, our approach employs counterfactual values to assess individual contributions toward global objectives rather than neighbor-specific objectives, thus promoting more efficient collaborative strategies at the global/system level.
To better facilitate credit assignment during the optimization of cooperative objectives, we design a new dual-level interaction-aware centralized critic that effectively identifies both local and global interactions among vehicles.
Specifically, we use variational inference techniques to derive latent variables representing the pairwise interactions between agents by predicting the impact of each target agent on the ego agent's observation-state transition and reward signals.
We then integrate these interaction-aware latent variables into the global state-action information of all other vehicles and utilize an attention mechanism from the graph attention network (GAT) to capture global interactions and dependencies.
By modeling both micro- and macro-level agent interactions, COIN enables more accurate global value estimation and more reliable policy learning in dense, highly interactive multi-agent navigation scenarios.

We conduct simulation experiments in urban environments using the open-source MetaDrive~\cite{li2022metadrive} simulator, which is designed for autonomous driving.
We evaluate COIN against advanced learning-based methods in diverse traffic scenarios, including intersections and roundabouts, and bottlenecks.
There, our empirical results demonstrate that COIN outperforms baseline methods in terms of efficiency and safety under different system sizes, highlighting its effectiveness in collaborative navigation for MASD systems.
The performance gains arise from our CIG-TD3 algorithm, which jointly optimizes navigation and cooperation objectives, enabling agents to efficiently achieve their goals while maintaining coordination, and from our interaction-aware centralized critic, which enhances global value estimation and credit assignment for improved collaborative policy optimization.
Additionally, we experimentally validate learned trajectories on real mobile robots in an hybrid urban intersection mockup, demonstrating COIN's effectiveness and adaptability towards real-world applications.

In summary, the main contributions of this work are as follows:
\begin{enumerate}
    \item We propose COIN, a collaborative and interaction-aware MARL framework for MASD systems in densely interactive traffic environments, which explicitly formulates continuous navigation control as the joint optimization of individual driving objectives and global cooperative objectives.
    \item Within the COIN framework, we develop the CIG-TD3 algorithm, which simultaneously trains a local critic for ego-vehicle navigation and a centralized critic for system-level coordination. It further leverages counterfactual values to assess each agent’s contribution to the global objectives, enabling more accurate credit assignment and effective learning of collaborative navigation strategies.
    \item To improve value estimate during the optimization of global objectives, we design a dual-level interaction-aware centralized critic. It captures micro-level pairwise interactions using variational inference and models macro-level global interactions through an graph attention mechanism.
    \item We evaluate COIN against other advanced RL methods across three dense and highly interactive traffic scenarios, including intersections, roundabouts, and bottlenecks, under different system sizes, showing that it outperforms the baselines in safety, navigation efficiency, and overall performance.
    Finally, we deploy the learned policy on real robots in a physical urban mockup, demonstrating its effectiveness and adaptability in realistic MASD environments.
    
\end{enumerate}

The remainder of this paper is organized as follows: Section~\ref{sec:related_work} reviews the related work. 
Section~\ref{sec:background} introduces the MARL formulation of the navigation problem in MASD systems. 
Section~\ref{sec:method} describes the proposed COIN framework, a collaborative and interaction-aware MARL approach.
Section~\ref{sec:experiments} presents extensive experimental evaluations and discusses the results. 
Finally, Section~\ref{sec:conclusion} concludes the paper with a summary of the findings and outlines potential future research directions.

\section{Related Works}
\label{sec:related_work}

\subsection{Multi-Agent Reinforcement Learning}
\label{sec:related_work_1}

Collaborative MARL algorithms have been playing a crucial role in enabling effective cooperation and coordination in multi-agent systems, where agents must work together to achieve shared objectives in complex, dynamic environments.
The CTDE paradigm has become popular in collaborative MARL due to its ability to leverage global information (such as joint states and joint actions) during training and utilize local information (e.g., observations) during execution, which effectively combines the strengths of both centralized and decentralized methods. 
One representative method within this framework is MADDPG~\cite{lowe2017multi}, which extends the deep deterministic policy gradient (DDPG) algorithm to multi-agent settings by equipping each agent with a decentralized actor and a centralized critic conditioned on the joint observations and actions. 
This design allows agents to learn robust policies in continuous control environments while enabling coordination through shared training signals.
Another notable CTDE method was Value Decomposition Networks (VDN)~\cite{sunehag2017value}, which simplified multi-agent value function learning by decomposing the joint state-action-value function into a summation of individual value functions.
QMIX~\cite{rashid2020monotonic} extended this idea by employing a mixing network that combines individual value functions under a monotonicity constraint, enabling a more expressive joint state-action-value representation while preserving consistency between local and global optima.
COMA~\cite{foerster2018counterfactual} addressed the multi-agent credit assignment problem under the CTDE paradigm by introducing a counterfactual baseline, which marginalizes out an agent’s action to better estimate its individual contribution to the team performance.
More recent CTDE-based methods have further explored challenges such as scalability and training stability, which are particularly critical for complex and large-scale multi-agent systems.
For example, MAAC~\cite{iqbal2019actor} improved scalability in complex environments by incorporating attention mechanisms into the centralized critic, allowing each agent to selectively attend to relevant others based on context, especially when the number of agents varies.
MAPPO~\cite{yu2022surprising} extended Proximal Policy Optimization (PPO) to the multi-agent domain by using a shared centralized critic network for training, which stabilized learning and led to strong empirical performance in a variety of cooperative benchmarks.

\subsection{MARL for Self-Driving Systems}
\label{sec:related_work_3}
MARL has been widely applied in multi-robot systems for tasks such as autonomous exploration~\cite{tan2024ir, wang2024viper}, path planning~\cite{damani2021primal, qi2024bidirectional, he2025social}, and mapless navigation~\cite{zhang2022pipo, yang2023learning}, as well as in intelligent transportation systems for traffic signal control~\cite{chu2019multi,  han2023leveraging, zhang2025coordlight, zhang2025unicorn} and multi-vehicle control and coordination~\cite{yang2019cm3, palanisamy2020multi, peng2021connected, toghi2022social, han2023multi, zheng2024safe, cai2024interaction}. 
These methods leverage data-driven learning framework to tackle complex multi-agent decision-making problems in dynamic and partially observable environments, enabling agents to learn adaptive and coordinated behaviors.
However, most existing approaches focus on high-level decision-making in discrete action spaces, which simplifies policy learning but neglects the continuous and fine-grained control essential for end-to-end autonomous driving~\cite{zhang2024multi, hua2025multi}.
In these tasks, agents (vehicles) must learn to generate low-level continuous actions (such as steering and throttle) directly from raw observations, which presents significant challenges for the stability of policy learning and the overall coordination of the system.

MARL has been further explored for end-to-end navigation in MASD systems, with a particular focus on coordinating multiple AVs in structured and densely interactive traffic environments.
Previous research has focused on optimizing neighborhood performance by aligning agent objectives to improve collaboration and cooperation.
For instance, CoPO~\cite{peng2021learning} introduced Local Coordination Factors (LCFs) to balance each agent’s individual objective with those of its neighbors. It further proposed a bi-level meta-gradient learning approach that adjusts the LCFs based on a global objective, encouraging more cooperation and diverse social driving behaviors.
Similarly, SoLPO~\cite{chen2025mixed} proposed a mixed-motive MARL framework that combines individual and social learning, allowing agents to balance self-interest with socially coordinated behavior. It also introduced a social reward integration module that fuses individual and neighbor rewards to enhance coordination and learning efficiency.
TraCo~\cite{liu2023traco}, inspired by traffic coordinators, combined cross-attention and counterfactual advantage functions to extract distinctive characteristics of domain agents and accurately quantify each agent's contribution.
Additionally, SAPO~\cite{dai2023socially} employed a self-attention module to identify the most interactive participants and a social-aware integration mechanism to align the ego vehicle’s objectives with its partners by updating their social preferences. 
Another line of research focused on enhancing implicit coordination through effective communication and message aggregation.
Self-supervised Message Attention Encoding (SMAE)~\cite{liang2024limited} addressed the challenge of partial observability by enabling vehicles to receive and aggregate limited messages from nearby vehicles via an attention mechanism.
Liang et al.~\cite{liang2024efficient} proposed MACAC, which used attention-based communication to enhance teammate modeling and agent collaboration for MASD systems.
Despite these advancements, existing methods often fail to fully capture the complex interactions among agents in dense and highly interactive traffic environments.
Moreover, they struggle to effectively optimize global objectives that may conflict with individual goals, leading to suboptimal coordination and reduced overall system performance.
Unlike prior MARL approaches such as TraCo and SAPO that focus mainly on neighborhood-level objective optimization or local interactions, COIN models both fine-grained and system-level dependencies through a centralized critic and jointly optimizes local and global objectives within a unified CTDE framework.

\section{Background}
\label{sec:background}

In this section, we formulate the vehicle navigation problem in MASD systems as a MARL problem and present the corresponding problem formulation.
Details of the RL agent design, including observation, action, and reward definitions, are provided in the appendix.

Due to the decentralized nature and limited observations in MASD systems, we formulate the multi-vehicle navigation problem as a MARL problem, which is modeled as a Decentralized Partially Observable Markov Decision Process (Dec-POMDP)~\cite{gupta2017cooperative}, represented by the tuple $\mathcal{G} = <\mathcal{N}, \mathcal{S}, \mathcal{A}, \mathcal{P}, \mathcal{R}, \rho_0, \mathcal{O}, \mathcal{Z}, \gamma>$.
Here, $\mathcal{N}$ is the set of agents (vehicles), and $\mathcal{N}_t=\left\{i_{1, t}, \ldots, i_{n_t, t}\right\} \subset \mathcal{N}$ denotes the set of $n_t$ agents active at time step $t$.  
$\mathcal{S}$ is the global state space, and $\mathcal{A}=\bigcup_{i=1}^{n_t} \mathcal{A}_{i, t}$ is the joint action space of all active agents.
We consider a partially observable setting that agents cannot directly access the environment's true state $s \in \mathcal{S}$; instead, each agent obtains individual observation $o_{i,t}$ through the observation function $\mathcal{Z}_i\left(s_t\right): \mathcal{S} \rightarrow \mathcal{O}$, such that $o_{i, t}=\mathcal{Z}_i(s_t)$, where $\mathcal{O}$ is the observation space. 
Each agent $i$ selects its action $a_{i, t} \in \mathcal{A}_{i, t}$ which contributes to a joint action $\mathbf{a}_t=\left\{a_{i, t}\right\}_{i=1}^{n_t} \in \mathcal{A}_t$. 
The execution of this joint action causes an environmental state transition via the transition function $\mathcal{P}\left(s_{t+1} \mid s_t, \mathbf{a}_t\right)$, following which each agent receives an individual reward determined by the reward function $r_{i, t}=\mathcal{R}_i\left(s_t, \mathbf{a}_t\right)$. 
$\rho_0$ and $\gamma$ are the initial state distribution and the discount factor, respectively.

Given the policy $\mu_i$ (i.e., a function that maps from observations to actions) parameterized by $\theta_i$, we define the \textbf{individual objective}(navigation) for each agent as the discounted sum of its own rewards: $J_i^I(\theta_i) = \mathbb{E}[\sum_{t=t_i^s}^{t_i^e} \gamma^{t-t_i^s} r_{i, t}]$, where $t_i^s$ and $t_i^e$ denote the time steps when agent $i$ enters and exits the environment, respectively. 
Moreover, we define the \textbf{global objective} (collaboration) for each agent as the cumulative average reward across all active agents at each time step: $J_i^G(\theta_i) ={\mathbb{E}}\left[\sum_{t=t_i^s}^{t_i^e} \frac{\sum_{j \in \mathcal{N}_{t}} r_{j, t}}{\left|\mathcal{N}_{t}\right|}\right]$. 
Thus, the total objective that each agent aims to optimize is: $J(\theta_i)=J_i^I(\theta_i)+J_i^G(\theta_i)$.

\section{Collaborative Interaction-aware MARL}
\label{sec:method}
In this section, we present COIN, a collaborative interaction-aware MARL framework for end-to-end navigation in dynamic MASD systems. 
We first introduce a novel centralized critic architecture that integrates both local and global interaction-aware modules to capture local pairwise agent interactions and global agent dependencies, respectively. 
Building on this architecture, we propose the CIG-TD3 algorithm, which jointly optimizes individual (navigation) and global (collaboration) objective.
By balancing the goals of vehicle navigation and multi-agent collaboration in MASD systems, this design enables more efficient and safer collaborative navigation in complex and dynamic traffic environments.

\begin{figure*}[t!]
    \centering
    \includegraphics[width=\textwidth, height=0.6\textwidth]{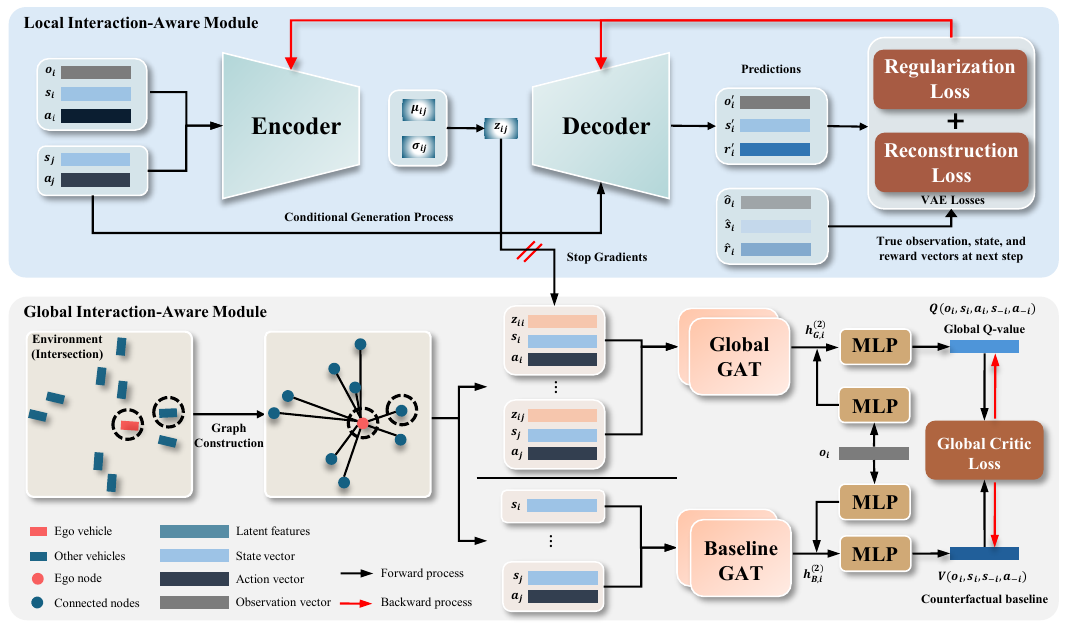}
    \caption{
    The structure of our proposed dual-level interaction-aware centralized critic network. It includes a local interaction-aware module to capture pairwise interactions between agents, and a global interaction-aware module to dynamically model global dependencies. These two modules work together to estimate the global Q-value. In addition, a counterfactual baseline (which is not conditioned on the ego agent’s action information) is used to improve credit assignment and guide cooperative policy learning during centralized training.
    }
    \label{fig:algorithm} 
\end{figure*}

\subsection{Interaction-aware Centralized Critic}
\label{sec:method_critic}
\subsubsection{Local Interaction-Aware Module}
Given the dynamics of the MASD systems, agents need to capture interactions to identify critical neighbors and make informed decisions.
The key interaction between two agents can be broadly summarized into two categories: (1) impact on observation/state transitions, which reflects how the presence and actions of one agent influence the state dynamics of the other, and (2) impact on received reward signals, indicating how one agent's behavior affects the rewards of the other.
Understanding these interactions is crucial for developing collaboration strategies and improving overall system performance.

To achieve this, we employ Conditional Variational AutoEncoders (CVAEs)~\cite{sohn2015learning} as the backbone of our local interaction-aware module, which leverage privileged global information, including each agent’s local state (a 10-dimensional vector in the world coordinate frame that consists of position, heading, velocity, and two navigation point vectors), along with the current observation and action, to reconstruct the ego agent's observation, local state, and reward at the next step. 
By conditioning on both the ego agent’s and the target agent’s information, this module models pairwise interactions in a unified manner, capturing their influence on both state transitions and received rewards, which correspond to the two types of interactions defined earlier.
The resulting latent variables provide a compact and structured representation of these interactions.
For a pair of agents $i$ (ego) and $j$ (considered), agent $i$’s current observation, local state, action, reward, next observation, and next local state are denoted as $o_i$, $s_i$, $a_i$, $r_i$, $o^{\prime}_{i}$, and $s^{\prime}_{i}$, respectively, while agent $j$’s current observation, local state, and action are $o_j$, $s_j$, and $a_j$.

As shown in the upper half of Fig.~\ref{fig:algorithm}, our CVAE model consists of an encoder and a decoder, parameterized by $\psi_e$ and $\psi_d$, respectively. 
The encoder takes as input the observation $o_i$, local states $s_i$ and $s_j$, and actions $a_i$ and $a_j$ from both the ego agent and its neighboring agents. 
It then maps these inputs to 20-dimensional mean and variance vectors, $\mu_{ij}$ and $\sigma_{ij}$, of a latent distribution, from which the latent variable $z_{ij}$ is sampled. 
Thus, the encoder that used to approximate the posterior is written as:
\begin{equation}
\textbf{Encoder}: q_{\psi_e}\left(z_{ij} \mid o_i, s_i, s_j, a_i, a_j\right).
\end{equation}
The decoder then reconstructs the concatenation of the next observation $o_i^{\prime}$, next local state $s_i^{\prime}$, and the reward $r_i$ received after taking action $a_i$, which is denoted as $y_{\text{vae}}=[o_i^{\prime}, s_i^{\prime}, r_i]$, from the latent variable $z_{ij}$ and the input vectors. 
The decoder is expressed as follows:
\begin{equation}
\textbf{Decoder}: p_{\psi_d}\left(y_{\text {vae}} \mid z_{ij}, o_i, s_i, s_j, a_i, a_j\right).
\end{equation}
The training objective of the CVAE is to maximize the Evidence Lower Bound (ELBO), which involves two key components: maximizing the log likelihood to ensure accurate data reconstruction and minimizing the Kullback-Leibler (KL) divergence between the approximate posterior and the prior distribution to regularize the latent space. The total ELBO objective is formulated as follows:
\begin{equation}
\label{cvae_loss}
\begin{aligned}
& \mathcal{L}\left(\psi_e, \psi_d\right)=-\underbrace{\operatorname{KL}\left(q_{\psi_e}\left(z_{ij} \mid o_i, s_i, s_j, a_i, a_j\right) \| p(z_{ij})\right)}_{\text {Regularization (KL. Divergence)}}+ \\
&\underbrace{\mathbb{E}_{q_{\psi_e}\left(z_{ij} \mid o_i, s_i, s_j, a_i, a_j\right)}\left[\log p_{\psi_d}\left(y_{\text {vae}} \mid z_{ij}, o_i, s_i, s_j, a_i, a_j\right)\right]}_{\text {Reconstruction Loss}}.
\end{aligned}
\end{equation}
Here, the first term, KL divergence, regularizes the latent space to align with a prior distribution, typically a standard normal distribution.
The second term represents the reconstruction log likelihood, ensuring that the decoder can accurately reconstruct the ego agent's next observation $o^{\prime}_i$, next state $s^{\prime}_i$, and the reward $r_i$.
The key to the reconstruction process lies in extracting crucial interaction information between agents through the latent variable $z_{ij}$.
By incorporating other agents' current states and actions into the reconstruction, the model captures their influence on the ego agent's future observations, state transitions and rewards.
This enables agents to learn a concise and informative representation of pairwise interactions in the latent space, which improves the estimation of global Q-values by explicitly incorporating micro-level interactions, thereby enhancing the collaboration performance in complex multi-agent scenarios.

\subsubsection{Global Interaction-Aware Module}

To model global interactions among agents in MASD systems, we adopt the attention mechanism from the Graph Attention Network (GAT)~\cite{velivckovic2017graph}, applied over a fully connected graph where each node corresponds to an agent and each edge models the potential interaction between a pair of agents. 
This enables the ego agent $i$ to focus on the most relevant neighboring agents $j \in \mathcal{N}$ and selectively aggregate their information to support accurate estimation of its contribution to the global objective. 
While we use a fully connected graph due to the relatively small number of interacting agents within local regions of dense traffic scenarios, the connectivity structure is flexible and supports adjustments (e.g., to sparse graphs) when applied to larger-scale environments.
For each neighboring agent $j$, the input $x_j$ is constructed by concatenating the local state $s_i$, local action $a_i$, and the interaction-conditioned latent variable $z_{ij}$ from the local interaction-aware module. 
Formally, this is represented as: $x_j=[s_i, a_i, z_{ij}], \; \forall j \in \mathcal{N}$.
The input of each agent is then processed through a shared Multi-Layer Perceptron (MLP), which transforms the concatenated input into a 128-dimensional feature vector. 
The MLP consists of two linear layers with LeakyReLU activation functions, and this transformation is described as: $h_j^{(0)}=\operatorname{MLP}\left(x_j\right), \; \forall j \in \mathcal{N}.$
The output vector, $h_j^{(0)}$, serves as the initial node features for the GAT. 
The GAT consists of two layers, where each layer computes attention coefficients and aggregates features from neighboring agents.
In each GAT layer, the attention mechanism computes a coefficient for each pair of connected agents to determine the importance of neighboring node features. 
The attention score $e_{ij}$ between agent $i$ and its neighbor agent $j$ is first computed as:
\begin{equation}
e^{(0)}_{i j}=\text { LeakyReLU }\left(\mathbf{a}_{\text{gat}}^T\left[\mathbf{W} h^{(0)}_i \| \mathbf{W} h^{(0)}_j\right]\right),
\end{equation}
where $\mathbf{W}$ is a learnable weight matrix, $\mathbf{a}_{\textit{gat}}$ is a learnable attention vector, and $\|$ denotes concatenation. 
The attention coefficient $\alpha_{i j}$ is then obtained by applying the softmax function over the attention scores:
\begin{equation}
\alpha_{i j}=\operatorname{softmax}_j\left(e_{i j}\right)=\frac{\exp \left(e_{i j}\right)}{\sum_{k \in \mathcal{N}_i} \exp \left(e_{i k}\right)}.
\end{equation}
The node features are then updated by aggregating the neighboring node features weighted by these attention coefficients, yielding the updated feature vector $h^{(1)}_i$:
\begin{equation}
h_i^{(1)}=\sigma\left(\sum_{j \in \mathcal{N}_i} \alpha_{i j}^{(0)} \mathbf{W} h_j^{(0)}\right).
\end{equation}
Here $\sigma$ denotes the ReLU activation function~\cite{agarap2018deep}.
The same GAT computation process is applied in the second GAT layer, which takes $h^{(1)}_i$ as input, computes and normalizes the attention scores, and aggregates the features from other nodes to produce the final feature vector $h^{(2)}_i$.
As shown in the lower half of Fig.~\ref{fig:algorithm}, the output vector $h^{(2)}_{G,i}$ from the global GAT is further concatenated with the feature vector generated from the observation vector through a two-layer MLP, followed by a one-layer MLP to output the global Q-value.
Additionally, we calculate a counterfactual baseline for the global Q-value using a similar process through a separate baseline GAT (i.e., through $h^{(2)}_{B,i}$).
The key difference is that it removes the ego agent’s actions and the latent variables from the input vectors.  
This counterfactual baseline is designed to estimate a global Q-value that excludes the influence of the ego agent, enabling the computation of its marginal contribution to the global collaborative objectives. This design follows the principle of counterfactual credit assignment and facilitates more accurate and disentangled evaluation of individual impact on system-level coordination.
By using the GAT to process both global state-action information and local interaction-aware features, our global interaction-aware module effectively captures inter-agent dependencies, thereby enhancing its ability to represent macro-level interactions and enhancing cooperative learning by providing more accurate value estimation and better guidance for policy updates..
In summary, our dual-level interaction-aware centralized critic combines a local and a global interaction-aware module to explicitly model pairwise agent interactions and system-wide dependencies. 
This design improves the accuracy of global Q-value and counterfactual baseline estimation, providing more reliable learning signals for optimizing global objectives in complex and dynamic multi-agent environments.

\subsection{Counterfactual Individual-Global TD3 (CIG-TD3)}
\label{sec:method_algorithm}

To jointly optimize both individual navigation and global collaboration objectives for continuous end-to-end navigation tasks in MASD systems, we develop a refined variant of the Twin Delayed Deep Deterministic Policy Gradient (TD3) algorithm~\cite{fujimoto2018addressing}, named counterfactual individual-global TD3 (CIG-TD3).
In addition to individual Q-values, this variant introduces global Q-values and their counterfactual baselines, estimated by the interaction-aware centralized critic described in Sec.~\ref{sec:method_critic}.
CIG-TD3 follows the CTDE paradigm and adopts parameter sharing, allowing agents to update shared network parameters to improve learning efficiency and stability.
Although additional global Q-values are introduced during training, they are used only to guide policy optimization.
Specifically, the local critic estimates Q-values to guide policy learning toward individual objectives, while the global centralized critic leverages both local observations and privileged global information to guide policy learning toward global objectives.
At execution time, each agent selects actions based solely on its local observation through the actor network, ensuring a lightweight and scalable control framework for MASD systems.

\subsection{Critic Learning}
To optimize the local critic networks, we follow the original Q-learning structure from the vanilla TD3 algorithm~\cite{fujimoto2018addressing}, which employs double Q-learning and delayed policy updates to mitigate overestimation bias and improve training stability.
Given an experience buffer $\mathcal{D}$ that stores past trajectories, the two local critic networks, parameterized by $\phi_1^l$ and $\phi_2^l$, are trained to minimize the mean squared error (MSE) between the estimated individual Q-values and and their corresponding targets. 
For each sampled transition $(o_i, s_i, \bm{s_{-i}}, a_i, \bm{a_{-i}}, r^{I}_{i}, r^{G}_{i}, d_i, o^{\prime}_{i}, s^{\prime}_{i}, \bm{s_{-i}^{\prime}}, \bm{a_{-i}^{\prime}}) \in \mathcal{D}$, where $\bm{s_{-i}}$ and $\bm{a_{-i}}$ denote the current local states and actions of all agents excluding agent $i$, and $\bm{s^{\prime}_{-i}}$, $\bm{a^{\prime}_{-i}}$ denote their next-step counterparts, the loss for local critic networks is defined as follows:
\begin{equation}
\begin{aligned}
\label{local_critic_loss}
L\left(\{\phi_k^l\}_{k=1}^{2}, \mathcal{D}\right) = \sum_{k=1}^{2} \underset{{\mathcal{D}}}{\mathbb{E}} \left[\left(Q^I_{\phi_k^l}(o_i, a_i) - y^I_i\right)^2\right],
\end{aligned}
\end{equation}
where the target individual Q-value $y^I_i$ is computed using the smoothed target policy and the minimum value from the two target local critics, which is expressed as:
\begin{equation}
\label{local_critic_target}
y^I_i=r^I_i+\gamma(1-d_i) \min _{k=1,2} Q^I_{\phi^l_{k, \text {targ}}}\left(o^{\prime}_i, a^{\prime}_i\right),
\end{equation}
where $d_i \in \{0, 1\}$ indicates whether the next state is a terminal state. 
To compute the target action at next step $a_i^{\prime}$ and $\bm{a_{-i}'}$, we use the target policy $\mu_{\theta_{\text{targ}}}$ and add clipped Gaussian noise to each action dimension. 
This encourages exploration while keeping the action within a valid range. 
The target action is then computed as:
$a_i^{\prime} = \operatorname{clip}\left(\mu_{\theta_{\text {targ}}}(o_i^{\prime}) + \operatorname{clip}(\epsilon, -c, c),\ a_{\text{Low}},\ a_{\text{High}}\right)$,
where $\epsilon \sim \mathcal{N}(0, \sigma)$ is the added noise, $c$ is a fixed noise threshold, and $a_{\text{Low}}$, $a_{\text{High}}$ are the lower and upper bounds of the action space.

Additionally, the global centralized critic networks that estimate the discounted cumulative global rewards, parametrized by $\phi_1^g$ and $\phi_2^g$, are trained by minimizing the MSE between the estimated global Q-values and the target global Q-values. The loss for global critic networks is defined as follows:
\begin{equation}
\begin{aligned}
\label{global_critic_loss}
L\left(\{\phi_k^g\}_{k=1}^{2}, \mathcal{D}\right) = \sum_{k=1}^{2} \underset{\mathcal{D}}{\mathbb{E}} \left[\left(Q^G_{\phi_k^g}(o_i, a_i, s_i, \bm{s}_{-i}, \bm{a}_{-i}) - y^G_i\right)^2\right].
\end{aligned}
\end{equation}
We further introduce a counterfactual baseline value, estimated by the same centralized critic described in Sec.~\ref{sec:method_critic} and parameterized by $\phi^b$. 
This counterfactual baseline leverages global information but excludes the ego agent’s action. It represents a state-conditioned expected value based on the ego agent’s observation $o_i$, its state $s_i$, the states of other agents $\bm{s_{-i}}$, and their joint actions $\bm{a}_{{-i}}$. Thus, the loss function for this counterfactual baseline $V_{\phi^{b}}$ is defined as:
\begin{equation}
\label{counterfactual_value_loss}
    L\left(\phi^b, \mathcal{D}\right)=\underset{{\mathcal{D}}}{\mathrm{E}}\left[\left(V_{\phi^b}(o_i, s_i, \bm{s}_{-i}, \bm{a}_{-i}) - y^G_i\right)^2\right].
\end{equation}
Here, $y^G_i$ is the target global Q-value, which is calculated as:
\begin{equation}
\label{global_critic_target}
y^G_i = r^G_i+\gamma(1-d_i) \min _{k=1,2} Q^G_{\phi^g_{k, \text {targ}}}\left(o^{\prime}_i, a^{\prime}_i, {s}^{\prime}_i, \bm{s}^{\prime}_{-i}, \bm{a}^{\prime}_{-i}  \right),
\end{equation}
where the target actions of other agents $\bm{a}^{\prime}_{-i}$ are computed in the same way as $a_i^{\prime}$ by applying the target policy with clipped Gaussian noise and clipping the actions to the valid action range.

\subsection{Actor Learning}
The loss function for the actor network (i.e., the policy) is designed to maximize the expected return by updating the policy in the direction that increases the Q-values.
In navigation tasks, using only individual Q-values to guide policy updates (e.g., IPPO~\cite{schulman2017proximal}, ITD3~\cite{fujimoto2018addressing}) often leads agents to learn selfish, sub-optimal strategies, which is harmful for the overall system efficiency. 
Conversely, focusing solely on global Q-values (e.g., MADDPG~\cite{lowe2017multi}, QMIX~\cite{rashid2020monotonic}, and MAPPO~\cite{yu2022surprising}) introduces high variance and credit assignment issues, resulting in "lazy agents" that condition on others' superior actions to achieve high global rewards, thereby failing to facilitate stable training and effective collaboration.
Recent neighborhood-based critics (e.g., TraCo~\cite{liu2023traco}, SAPO~\cite{dai2023socially}) offer a trade-off between individual and global Q-values by estimating value estimations over a set of nearby agents, but they rely on predefined interaction ranges and cannot capture system-wide cooperative effects.
In our proposed CIG-TD3 algorithm, the actor network (i.e., the policy, parameterized by $\theta$) is updated by optimizing both individual and global Q-values.
The key insight is that maximizing the individual Q-values enables each agent to learn how to navigate effectively in dynamic traffic environments, while maximizing the global Q-values encourages agents to cooperate with other agents and improve overall system performance.
To better address the credit assignment problem in complex MASD systems, we introduce a counterfactual baseline (i.e., $V_{\phi^{b}}$) when optimizing the global Q-values.
This baseline explicitly models the influence of other agents by conditioning on their states and actions, enabling a more accurate evaluation of the agent’s own contribution towards the global objective.
Thus, the actor loss function is defined as:
\begin{equation}
\label{actor_loss}
\begin{aligned}
& L(\theta, \mathcal{D})=-\underset{\mathcal{D}}{\mathbb{E}}\left[\underbrace{Q_{\phi_1^l}^I\left(o_i, \mu_{\theta}\left(o_i\right)\right)}_{\textit{Individual Objective}}+ \left(\underbrace{Q_{\phi_1^g}^G\left(o_i, \mu_{\theta}\left(o_i\right), s_i, \bm{s}_{-i}, \bm{a}_{-i}\right) - V_{\phi^b}\left(o_i, s_i, \bm{s}_{-i}, \bm{a}_{-i}\right)}_{\textit{Global Objective with Credit Assignment}}\right)\right].
\end{aligned}
\end{equation}
This linear combination follows common practice in cooperative MARL and is effective because it combines local value optimization with counterfactual global corrections, thereby mitigating gradient conflicts during training.
Lastly, we apply soft updates to the target networks for the local critic, global critic, and actor networks (e.g., $\theta_{targ} \leftarrow \tau_{\text{soft}} \; \theta + (1-\tau_{\text{soft}}) \; \theta_{targ}$) to reduce variance and promote stable learning over time, where $\tau_{\text{soft}}$ is a small hyperparameter that controls update rate of the target networks.
The detailed implementation of the CIG-TD3 algorithm is presented in Algorithm~\ref{algorithm:CIG-TD3}.

\begin{table}[h]
\centering
\caption{Comparative results of COIN and other baselines in three traffic environments. We report the average and standard deviation for six evaluation metrics, 
with the best performance in bold and the second-best underlined.
}
\resizebox{\textwidth}{!}{
\begin{tabular}{c|cccccc}
\hline
\multirow{2}{*}{Method} & \multicolumn{6}{c}{Intersection environment with 30 initialized  agents}                                                                                     \\ \cline{2-7} 
                        & Success Rate $\uparrow$           & Off-Road Rate $\downarrow$             & Collision Rate $\downarrow$           & Efficiency $\uparrow$            & Safety $\uparrow$               & Average Travel Steps $\downarrow$             \\ \hline
IPPO                    & 69.96 (15.21)          & 8.01 (4.73)          & 18.00 (7.20)         & 42.70 (24.5)            & -22.87 (6.99)         & 588.59 (122.50)         \\
MFPO                    & 69.95 (16.54)          & 7.41 (5.18)          & 19.08 (7.92)         & 44.73 (27.64)          & -23.76 (7.34)         & 579.62 (142.27)         \\
CPPO                    & 65.68 (13.34)          & 8.59 (4.32)          & 24.39 (8.14)         & 38.01 (25.05)          & -35.07 (8.58)         & 570.42 (109.10)         \\
CoPO                    & \underline{75.20 (12.99)}    & \underline{5.86 (3.08)}    & \underline{16.35 (7.64)}         & 49.20 (20.85)          & \underline{-19.59 (7.29)}   & 567.22 (102.62)         \\
ITD3                    & 72.66 (9.08)           & 5.61 (3.57)         & 21.82 (7.43)         & \underline{59.58 (22.79)}   & -36.98 (13.66)       & \textbf{469.18 (65.43))} \\
TraCo                   & 73.08 (9.07)           & 7.52 (4.01)          & 19.21 (7.88)         & 57.06 (22.81)   & -32.86 (11.50)        & \underline{482.16 (70.32)}   \\
COIN (ours)                   & \textbf{88.78 (10.09)} & \textbf{4.13 (3.90)} & \textbf{3.94 (3.74)} & \textbf{68.64 (15.47)} & \textbf{-6.73 (4.72)}  & 509.60 (79.94)    \\ \hline
                        & \multicolumn{6}{c}{Roundabout environment with 40 initialized agents}                                                                                       \\ \hline
IPPO                    & 66.99 (9.61)           & 7.82 (4.16)          & 10.52 (4.26)         & 40.16 (14.03)          & -15.09 (5.54)         & 738.75 (57.92)          \\
MFPO                    & 66.95 (9.63)           & 8.12 (4.01)          & 13.32 (5.60)         & 39.27 (13.47)          & -18.45 (5.78)         & 736.59 (44.43)          \\
CPPO                    & 71.57 (8.01)           & 7.02 (2.89)          & 20.20 (8.12)         & 45.82 (15.64)          & -28.66 (9.51)         & 667.04 (47.11)          \\
CoPO                    & 69.03 (12.17)          & \underline{5.16 (2.59)}    & \underline{11.82 (5.32)}   & 43.17 (15.72)          & \underline{-13.87 (5.13)}   & 729.40 (57.01)          \\
ITD3                    & \underline{81.12 (7.73)}     & 4.66 (2.95) & 12.14 (6.85)         & \underline{64.85 (15.73)}    & -16.94 (6.66)         & \underline{624.35 (63.41)}    \\
TraCo                   & 72.10 (6.22)           & 12.82 (5.98)         & 13.94(7.01)         & 48.93 (12.21)           & -29.38 (8.11)         & 649.67 (42.81)          \\
COIN (ours)                   & \textbf{90.68 (4.82)}  & \textbf{5.36 (2.99)}          & \textbf{3.22 (3.38)} & \textbf{80.66 (9.46)}  & \textbf{-8.58 (5.03)} & \textbf{579.01 (39.14)} \\ \hline
                        & \multicolumn{6}{c}{Bottleneck environment with 20 initialized agents}                                                                                       \\ \hline
IPPO                    & 71.46 (12.50)           & 17.70 (10.49)          & 11.10 (6.14)         & 34.12 (19.28)          & -22.16 (8.55)         & 523.20 (105.81)          \\
MFPO                    & 66.71 (12.43)           & 18.72 (8.67)           & 14.90 (6.98)         & 28.26 (21.70)          & -29.35 (11.91)        & 534.84 (94.61)         \\
CPPO                    & 70.08 (9.46)            & 15.99 (6.05)           & 14.16 (6.22)         & 33.68 (14.77)          & -26.61 (9.92)         & 512.12 (63.10)          \\
CoPO                    & 70.61 (11.23)           & 19.14 (9.17)           & 10.57 (5.21)         & 30.83 (17.27)          & -22.86 (9.54)         & 535.51 (78.55)          \\
ITD3                    & \underline{87.87 (5.69)}            & \underline{5.76 (2.71)}       & 6.63 (4.75)     & \underline{84.40 (13.33)}  & \underline{-13.87 (6.58)}     & \underline{301.27 (49.6)}    \\
TraCo                   & 86.44 (12.86)         & 7.90 (8.43)    &  \underline{6.12 (5.36)}   & 84.22 (30.18)          & -16.44 (15.31)    & 303.77 (107.50)         \\
COIN (ours)              & \textbf{96.33 (3.12)}   & \textbf{2.13 (1.77)}          & \textbf{1.57 (2.08)} & \textbf{96.91 (10.64)}  & \textbf{-3.79 (3.17)} & \textbf{246.42 (36.40)} \\ 
\hline
\end{tabular}}
\label{table:results_overall}
\end{table}

\begin{figure}
    \centering
    \includegraphics[width=\linewidth]{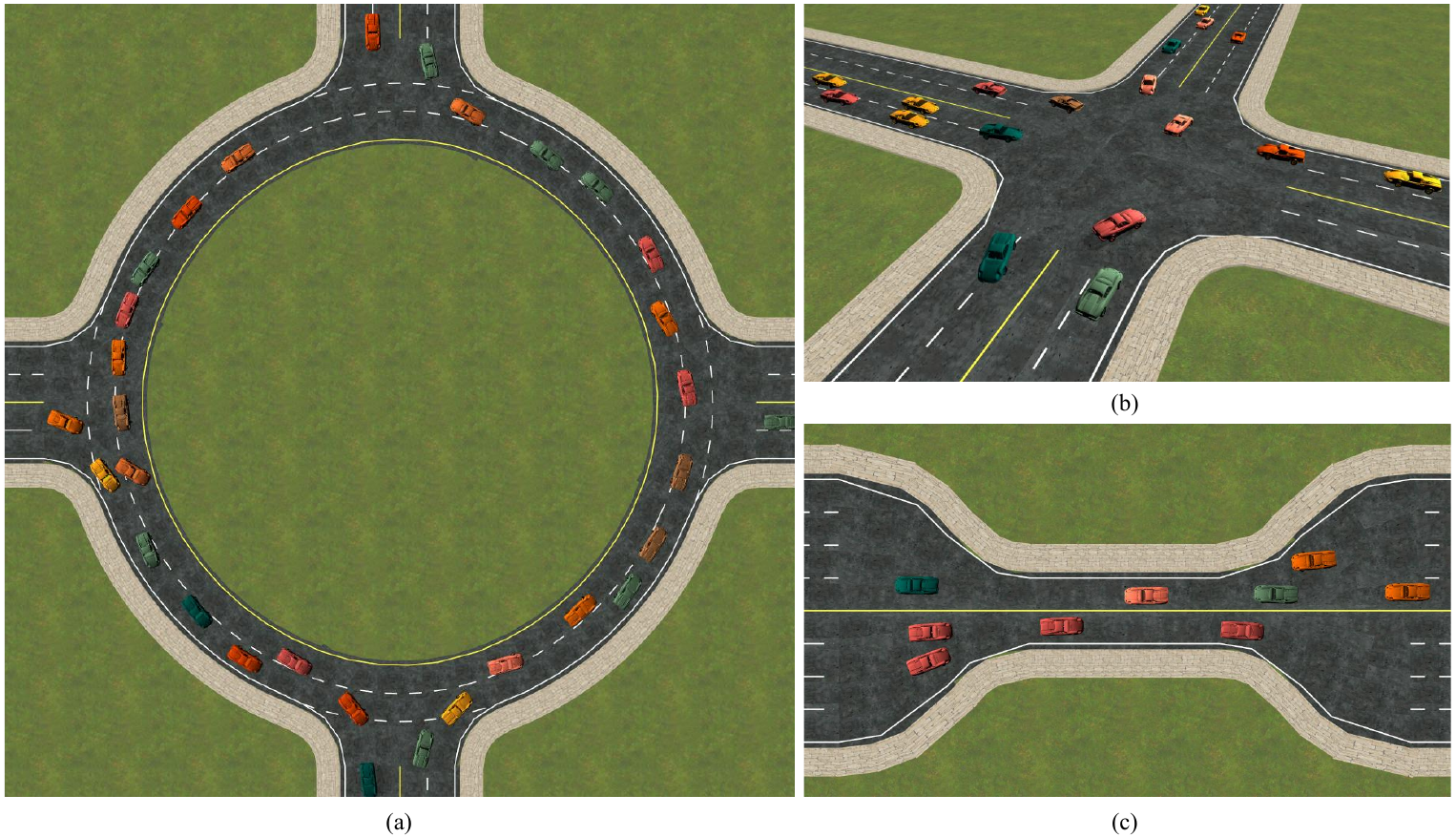}
    \caption{
    Illustration of three typical highly interactive and dense traffic scenarios for MASD systems in the MetaDrive simulator: (a) roundabout (left), (b) intersection (up right), and (c) bottleneck (bottom right).
    }
    \label{fig:dense_env}
    \vfill
\end{figure}

\begin{spacing}{0.9}
\begin{algorithm}
\caption{CIG-TD3 Algorithm}
\begin{algorithmic}[1] 
\State \textbf{Initialize:} actor network $\mu_\theta$ and its target; local critic network (outputting $Q_{\phi_1^l}, Q_{\phi_2^l}$) and its target for estimating individual Q-values; centralized critic network (producing $Q_{\phi_1^g}, Q_{\phi_2^g}$ and $V_{\phi^b}$) and its target for estimating global Q-values and counterfactual baselines; replay buffer $\mathcal{D}$
\For{step = 1 to step\_max}
    \State Observe joint state $\bm{s}$ and local observations $\{o_1, \dots, o_N\}$
    \For{each agent $i = 1$ to $N$}
        \If{step $<$ step\_start}
            \State Sample action $a_i \sim \text{Uniform}(\mathcal{A})$
        \Else
            \State Select action $a_i = \mu_\theta(o_i) + \epsilon$
        \EndIf
    \EndFor
    \State Execute joint action $\bm{a} = \{a_1, \dots, a_N\}$ in the environment
    \State Receive $\{r_1^I, \dots, r_N^I\}$, $\{r_1^G, \dots, r_N^G\}$, next joint state $\bm{s'}$, $\{o_1', \dots, o_N'\}$, and $\{d_1, \dots, d_N\}$
    \For{each agent $i = 1$ to $N$}
        \State Calculate next action $a_i' = \mu_\theta(o_i') + \epsilon$ 
    \EndFor
    \State Store $(o_i, s_i, \bm{s_{-i}}, a_i, \bm{a_{-i}}, r_i^I, r_i^G, d_i, o_i', s_i', \bm{s_{-i}'}, \bm{a_{-i}'})$ for all $i$ in buffer $\mathcal{D}$
    
    \If{length($\mathcal{D}$) $>$ batch\_size}
        \For{each agent $i = 1$ to $N$}
            \State Sample mini-batch for agent $i$ from buffer $\mathcal{D}$
            \State Update $Q_{\phi_1^l}, Q_{\phi_2^l}$ using Eqn~(7), with target values from Eqn~(8)
            \State Update $Q_{\phi_1^g}, Q_{\phi_2^g}$ and $V_{\phi^b}$ using Eqns~(9)--(11), jointly optimizing the CVAE in Eqn~(3)
            \If{step mod policy\_delay = 0}
            \State Update actor $\mu_\theta$ using Eqn~(12) based on $Q_{\phi_1^l}$, $Q_{\phi_1^g}$, and $Q_{\phi^b}$
                \State Soft update all target networks
            \EndIf
        \EndFor
    \EndIf
\EndFor
\end{algorithmic}
\label{algorithm:CIG-TD3}
\end{algorithm}
\end{spacing}

\section{Experiments and Results}
\label{sec:experiments}
In this section, we present a comprehensive evaluation of our proposed method. 
We begin by introducing the experimental setup, including simulated environments and evaluation metrics. 
We then describe the baseline methods used for comparison, followed by performance results covering overall effectiveness, generalization ability, and policy visualization. 
To further validate the contributions of key components, we conduct ablation studies. Finally, we demonstrate the real-world applicability of our method through experiments on a physical mobile robot system.

\subsection{Experimental Setup}

We conduct simulation experiments to compare COIN to various baselines using the open-source autonomous driving simulator MetaDrive~\cite{li2022metadrive}. 
The simulated environments are illustrated in Fig.~\ref{fig:dense_env}, with the descriptions of three dense and highly interactive traffic environments detailed as follows:
\begin{itemize}
    \item \textbf{Intersection}: a signal-free four-way intersection with bi-directional traffic. A total of 30 vehicles are initialized and respawned at random points on each road after reaching their destinations.
    \item \textbf{Roundabout}: a circular road layout where traffic flows around a central island. Roads connect to the circle from four directions, allowing vehicles to enter, circulate, and exit. We initialize 40 vehicles and allow them to respawn upon reaching their destinations.
    \item \textbf{Bottleneck}: a two-way road where each direction narrows from four lanes to one lane. Vehicles must merge and coordinate within the narrowing lane segment to avoid congestion. We initialize 15 vehicles, which will respawn after reaching their destinations.
\end{itemize}

To evaluate the performance of different methods, we adopt six evaluation metrics following the prior work CoPO~\cite{peng2021learning}, which include: \textbf{Success Rate}, the ratio of vehicles that reach their destinations to the total number of vehicles in an episode; \textbf{Collision Rate}, the ratio of vehicles involved in collisions with other vehicles to the total vehicle count; \textbf{Off-Road Rate}, the ratio of vehicles that drive off the road to the total number of vehicles; \textbf{Safety}, the negative total number of failures (e.g., crashes or off-road incidents) experienced by all vehicles in the system; 
\textbf{Efficiency}\footnote{All methods are evaluated under the same fixed time horizon, e.g., 1000 steps. The division by this constant is omitted in the reported values, as it does not affect relative comparisons.}, defined as a system-level outcome measured by the difference between the numbers of successful and failed vehicles within a fixed time horizon, following CoPO;
and \textbf{Average Travel Steps}, which measures the average number of steps taken by vehicles that successfully reach their destinations in an episode.

\subsection{Baseline Methods}
We compare COIN to various learning-based methods and its ablation variants, detailed as below:
\begin{itemize}
    \item \textbf{IPPO}: An independent learning version of the Proximal Policy Optimization (PPO)~\cite{schulman2017proximal} algorithm for multi-agent environments. Agents share a common policy network but update independently, treating other agents as part of the environment without explicit collaboration.
    \item \textbf{ITD3}: Similar to IPPO, but utilizes the twin
    delayed deep deterministic policy gradient (TD3) algorithm~\cite{fujimoto2018addressing} with parameter sharing for policy learning.
    \item \textbf{CPPO}: Extends IPPO with a centralized critic that aggregates all agents' information for a more comprehensive value estimate to optimize cooperation for improved system performance.
    \item \textbf{MFPO}: Combines IPPO with mean field MARL algorithms~\cite{yang2018mean}, where the centralized critic takes as input the average states of neighboring agents. 
    Here, mean field approximation averages the states of neighboring agents in the value estimate (i.e., centralized critic). This simplifies the modeling of interactions by reducing the complexity of accounting for each agent individually, while still capturing the overall influence of neighboring agents.
    \item \textbf{CoPO}~\cite{peng2021learning}: An advanced MARL algorithm that promotes bi-level coordination in MASD systems by introducing LCFs to balance individual and neighboring goals. CoPO optimizes these LCFs using a meta-gradient approach guided by a global system objective, enabling agents to learn socially-aware behaviors that align local incentives with overall cooperation.
    \item \textbf{TraCo}~\cite{liu2023traco}: Utilizes a cross-attention mechanism~\cite{vaswani2017attention} to capture crucial information from nearby vehicles and employs a counterfactual advantage function for credit assignment within a neighborhood range of 40 meters to improve agent collaboration.
\end{itemize}
All experiments are conducted on an Ubuntu server with 128 GB RAM, an Nvidia GeForce RTX 3090 GPU, and an AMD Ryzen 9 5950X 16-core processor.
We train each method 8 times with different random seeds (consistent across methods for fair comparison) to mitigate the impact of randomness. After each training, we test using 20 fixed random seeds (different from those used in training), which results in a total of 160 test episodes for each method.
For CoPO and its related baselines (MFPO, CPPO, IPPO), we follow their official implementations and use the hyperparameters provided in their codebases~\footnote{\url{https://github.com/decisionforce/CoPO}} to ensure fair and consistent comparisons.
For TD3-related methods, including ITD3, TraCo and COIN, we train for approximately one million environment steps and use hyperparameters consistent with those commonly adopted in TD3-based approaches.
We use a Gaussian exploration noise standard deviation of $0.05$, a batch size of $256$ for both actor and critic, a discount factor of $0.99$, and a target network update rate of $0.005$. The learning rates for the actor and critic networks are set to $2\times10^{-4}$ and $3\times10^{-4}$, respectively. Additionally, we add a noise of $0.1$ to the target policy during critic updates, clip this noise within a range of $0.5$, and apply delayed policy updates with a frequency of $5$. The hidden dimension/number of units for the neural networks is set to $256$.
Our codes will be released upon the paper acceptance.

We also analyze the computational cost of the proposed interaction-aware centralized critic.
The main overhead comes from the VAE and the GAT layers used to capture pairwise and system-level interactions.
The dominant model-side cost per update can be approximated as $O\left(B\left(N H^2+N^2 H+N D_{\mathrm{vae}} H\right)\right)$, where $B$ is the batch size, $N$ is the number of interacting agents, $H$ is the hidden dimension, and $D_{\mathrm{vae}}$ is the VAE input dimension.
Additionally, the actor is a three-MLP network with training complexity $O\left(B H^2\right)$, and the local critic follows the standard twin-critic structure in TD3 with the same order of complexity.
In our MASD settings, $N$ remains relatively small, so this additional cost stays within a manageable range.
Under the CTDE paradigm, both critics are used only during training, while execution relies solely on the actor ($\approx$ 0.26 ms per step), resulting in inference cost comparable to standard TD3-style methods.

\subsection{Results and Discussions}
\subsubsection{Overall Performance}

We evaluate COIN against various baselines in three typical dense and highly interactive traffic scenarios: intersections, roundabouts, and bottlenecks (where intersections have the highest complexity, followed by roundabouts, and bottlenecks are relatively simpler).
As shown in Table~\ref{table:results_overall}, COIN significantly outperforms all baselines in terms of success rate and collision rate.
We also perform paired t-tests between COIN and the advanced baselines (CoPO and TraCo) across three environments (intersection, roundabout, and bottleneck) and three key metrics (safety, efficiency, and success rate), resulting in 18 tests in total, all of which yield p-values lower than $1.67\times10^{-15}$. Using a Bonferroni correction for multiple tests, the final significance threshold is $p=5.56\times10^{-4}$ (based on a standard original threshold of $p=0.01$), indicating that the performance gains of COIN are statistically significant.
In the intersection environment, COIN achieves the highest success rate of 88.78\% and the lowest collision rate of 3.94\%, outperforming the second-best method (CoPO) by 13.58\% in success and 12.41\% in collision reduction.
In the roundabout environment, COIN also leads with a success rate of 90.68\% and a collision rate of 3.22\%, with improvements of 9.56\% and 8.6\% compared to the second-best methods (ITD3 and CoPO).
In the bottleneck environment, COIN reaches a success rate of 96.33\% and a collision rate of 1.57\%, with improvements of 9.56\% and 8.6\% compared to the second-best methods (ITD3 and CoPO).
These results demonstrate COIN's effectiveness and robustness across diverse and challenging traffic scenarios.

Furthermore, compared to independent learning methods such as IPPO and ITD3, COIN achieves markedly improved performance in both safety and efficiency. 
In the most challenging intersection environment, COIN enhances efficiency by 60.75\% over IPPO and by 15.21\% over ITD3. 
In terms of safety, COIN reduces the safety cost by 70.57\% compared to IPPO and by 81.80\% compared to ITD3.
These gains highlight COIN’s improved capability to promote inter-agent collaboration, resulting in more efficient and safer cooperative behavior in MASD systems.
Although ITD3 achieves slightly better performance in terms of average travel steps, its higher collision rate and lower success rate limit its overall effectiveness.
In contrast, COIN jointly optimizes individual and global objectives and uses counterfactual values to precisely assess each agent’s contribution to the global objective, addressing the high variance problem in credit assignment and enabling more efficient collaborative strategies in dense and highly interactive traffic environments.
Moreover, compared to CTDE-based methods such as CPPO, MFPO, CoPO, and TraCo, COIN still achieves leading performance across all key metrics.
Specifically, in the intersection environment, COIN improves efficiency by at least 20.29\% and reduces the safety cost by at least 65.65\%.
In the roundabout environment, COIN achieves a minimum improvement of 76.84\% in efficiency and 38.14\% in safety.
In the bottleneck environment, COIN further enhances efficiency by at least 15.17\% and reduces safety cost by at least 76.95\%.
This is because our design of a dual-level interaction-aware centralized critic, which models micro-level pairwise interactions using CVAE and captures macro-level global dependencies with a graph attention mechanism (GAT), further strengthens the model’s ability to understand complex global traffic states and to learn effective collaborative policies.
In summary, COIN demonstrates stable and significant advantages over both independent and cooperative baselines, validating its effectiveness and broad applicability in complex MASD systems.

\begin{figure*}[t]
    \centering
    \includegraphics[width=\linewidth]{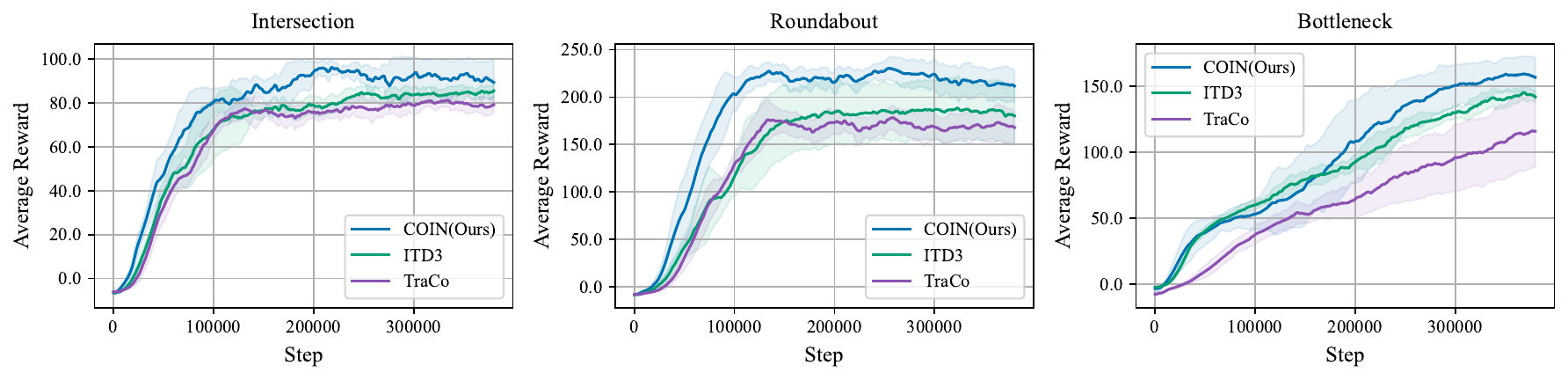}
    \caption{
    Smoothed average reward curves over training steps for COIN and baselines (e.g., ITD3 and TraCo) across three environments, where COIN consistently achieves higher rewards and faster convergence than other baselines.
    }
    \label{fig:training curve}
\end{figure*}

\begin{figure}[t]
    \centering
    \includegraphics[width=\linewidth]{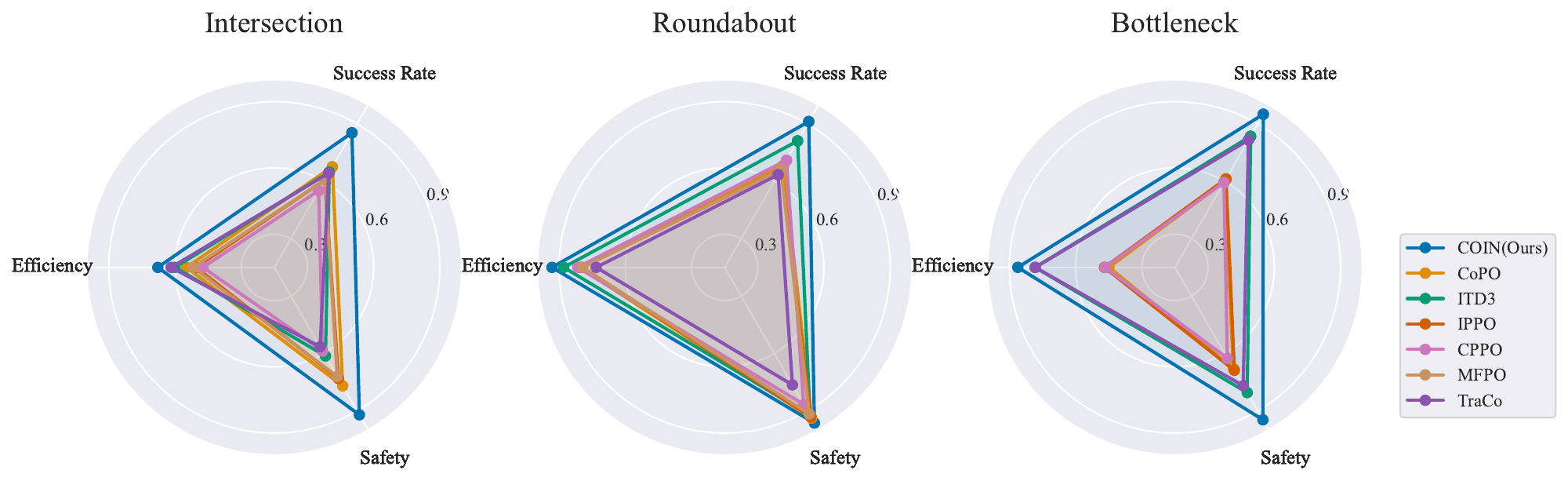}
    \caption{
    Performance comparison of COIN and other baselines across success rate, efficiency, and safety in three traffic environments, with radar plots showing that COIN forms the largest enclosed area in all scenarios.
    }
    \label{fig:results_radar}
\end{figure}

In addition, Figure~\ref{fig:training curve} presents the training curves, which showcase how the average reward evolves over time steps for COIN and selected TD3-based baselines (e.g., ITD3 and TraCo) across the three traffic environments.
To further visualize the performance difference between various baseline methods, we normalize the data for three key metrics, including success rate, safety, and efficiency, into a range of 0 to 1 and plot them in the radar chart in Fig.~\ref{fig:results_radar}.
This chart shows that COIN outperforms all baselines across these metrics, forming the largest area with the highest success rate, safety, and efficiency, highlighting its significant performance improvement in efficiency and safety.

\begin{figure}[t]
    \centering
    \includegraphics[width=\linewidth]{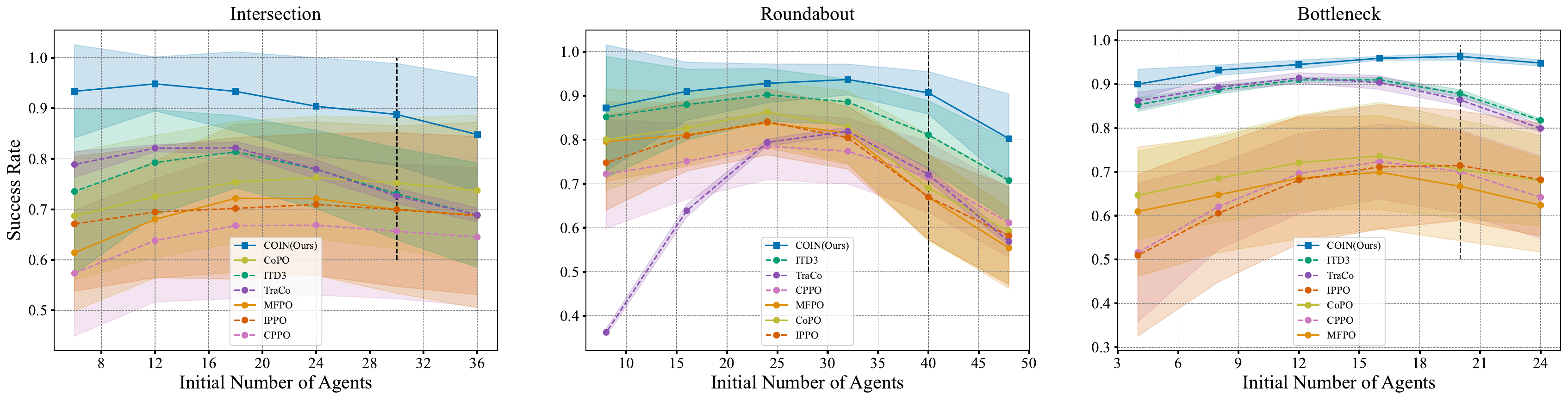}
    \caption{
    Generalization performance of various methods based on success rate. All methods are trained using the default initial agent counts, and tested with different counts to evaluate the zero-shot generalization ability. The black dashed line indicates the default initial agents counts during training, and in all scenarios and across all tested agent counts, COIN consistently achieves the highest success rate compared with other baseline methods.
    }
    \label{fig:results_generalization}
\end{figure}

\begin{figure*}[t]
    \centering
    \includegraphics[width=0.8\linewidth]{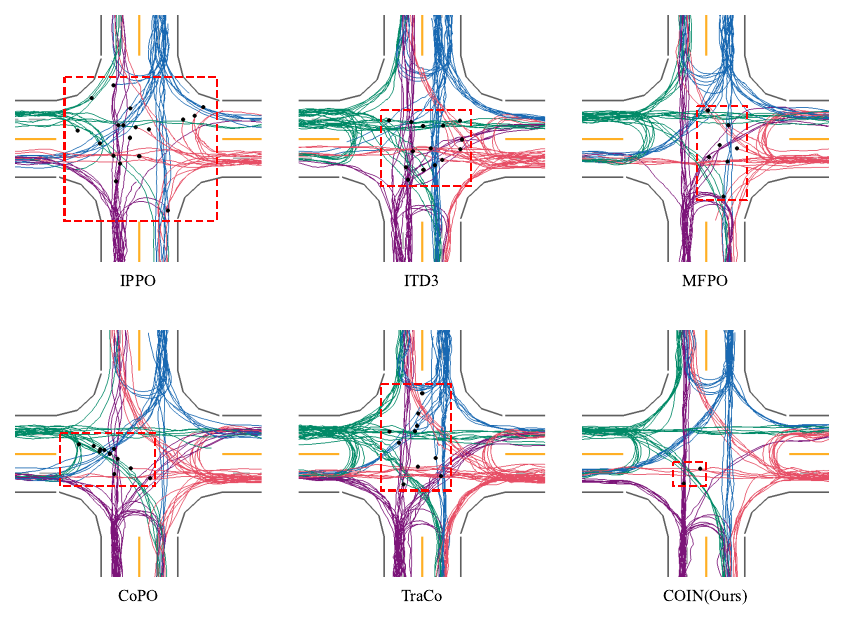}
    \caption{
    Trajectory visualizations of learned policies of various methods at the intersection environment, showing that COIN produces smoother and more consistent trajectories with fewer collisions compared to other methods.
    }
    \label{fig:results_traj_vis_intersection}
\end{figure*}

\begin{figure*}[t]
    \centering
    \includegraphics[width=0.9\linewidth]{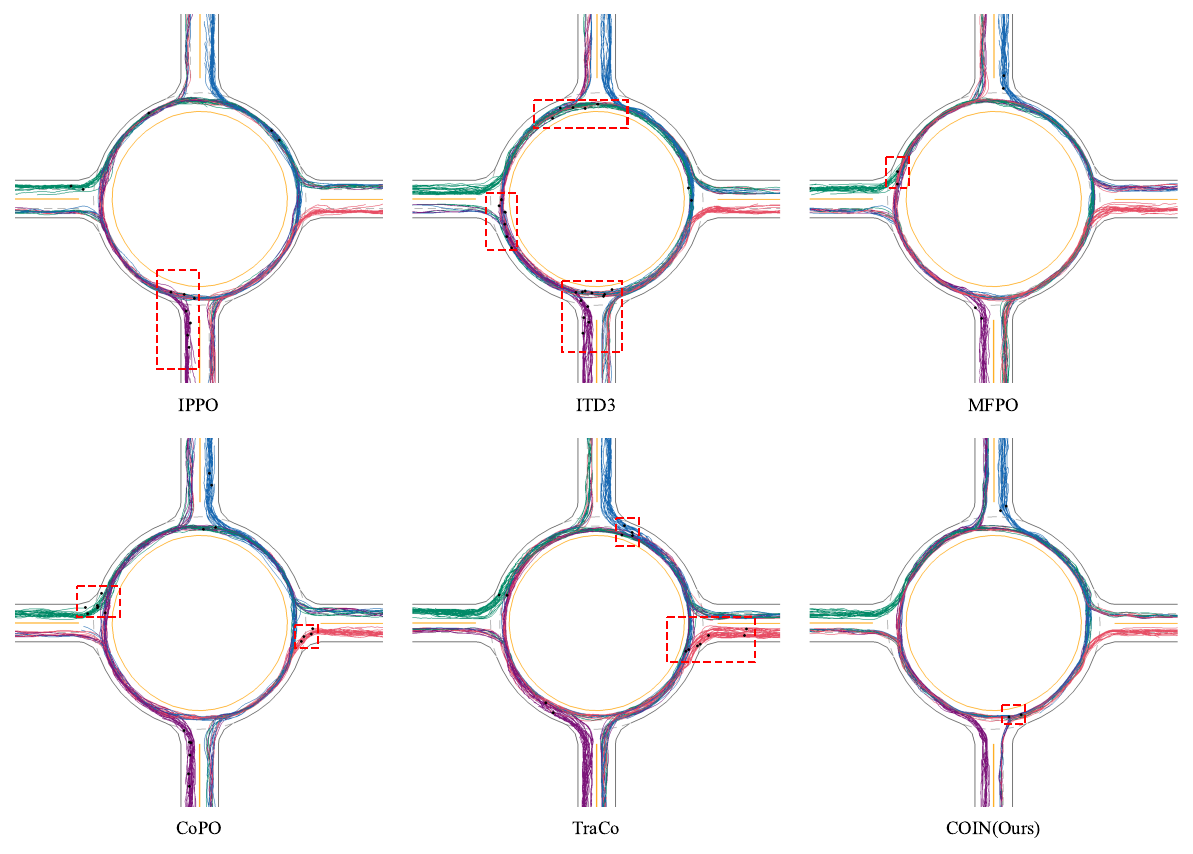}
    \caption{
    Trajectory visualizations of learned policies of various methods at the roundabout environment, demonstrating that COIN results in the fewest collisions among all methods.
    }
    \label{fig:results_traj_vis_roundabout}
\end{figure*}

\begin{figure*}[t]
    \centering
    \includegraphics[width=0.9\linewidth]{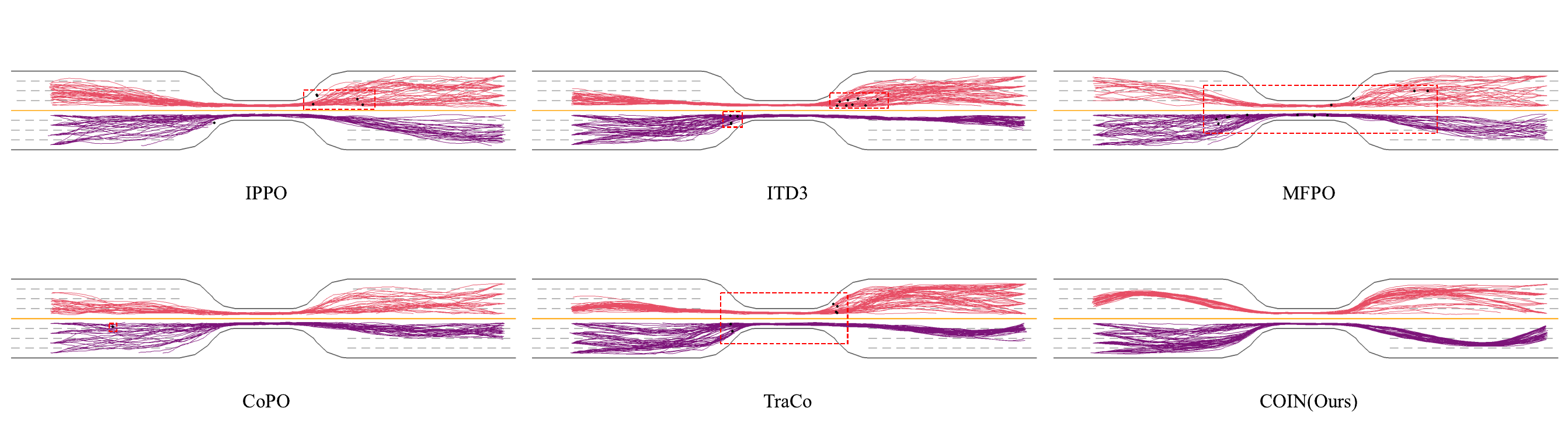}
    \caption{
    Trajectory visualizations of learned policies of various methods at the bottleneck environment, showing that COIN yields fewer collisions and smoother trajectory curves compared with other baseline methods.
    }
    \label{fig:results_traj_vis_bottleneck}
\end{figure*}

\subsubsection{Generalization Performance}
To evaluate generalization performance, we train all methods with default settings and test them with varying initial agent counts to assess their ability to generalize without retraining.
Fig.~\ref{fig:results_generalization} shows the success rate of each method running in the two scenarios with initial agent counts ranging from 6 to 36 in intersections and from 8 to 48 in roundabouts. 
In intersections, COIN achieves the highest success rate, remaining above 90\% when the number of initially generated vehicles is below 24. As this number increases, the success rate drops to 84\% with 36 vehicles, but still 10\% higher than the second-best method. 
CoPO follows with around 75\% success rate, while CPPO performs the worst overall. 
In roundabouts, COIN reaches nearly 100\% success with fewer vehicles, though its success rate declines slightly as vehicle numbers rise. 
ITD3 starts with stable performance but drops significantly at 32 vehicles, a trend also observed in the other baselines.
Overall, COIN demonstrates strong robustness and generalization, consistently achieving the highest success rates across various scenarios and system sizes, showing its superiority in handling dynamic MASD systems.

\subsubsection{Policy Visualization}

We select one test episode from each of the intersection, roundabout, and bottleneck environments and visualize the vehicle trajectories to visualize the learned policies.
The color of each trajectory represents vehicle's spawn roads, and the collision points are marked with black dots. Areas with frequent collisions are highlighted using red dashed boxes.
As shown in Fig.~\ref{fig:results_traj_vis_intersection}, IPPO exhibits the most collisions, with black dots scattered throughout the central area. 
This is because IPPO focuses solely on individual objectives, making it difficult to handle dense traffic scenarios, where agent cooperation is crucial.
MFPO and CoPO show fewer collisions, mostly concentrated in specific areas, suggesting that while these methods are able to handle relatively simple interactions, they have limitations in more complex, multi-directional traffic scenarios.
In contrast, COIN exhibits superior performance with fewest collisions and more consistent trajectories, highlighting its effectiveness in coordinating vehicles and improving both safety and efficiency.
Similar patterns can also be observed in the roundabout and bottleneck scenarios (see Fig.\ref{fig:results_traj_vis_roundabout} and Fig.\ref{fig:results_traj_vis_bottleneck}), where COIN continues to outperform other baseline methods with fewer collisions and smoother traffic flow. 
This demonstrates its ability to adapt to different traffic conditions while maintaining stable and cooperative driving strategies across diverse environments.

At \href{https://marmotlab.github.io/COIN/}{\textbf{this link}}, we provide simulation demonstrations of the policies learned by different methods in three environments described above, to further illustrate their learned behaviors.
In the interaction-intensive intersection environment, COIN demonstrates superior control and collaboration capabilities, exhibiting almost no collisions and very few off-road incidents. By accurately modeling interactions among vehicles during the optimization of global objectives, COIN achieves efficient and rational yielding and overtaking behaviors, thus enhancing overall traffic efficiency. In contrast, ITD3 and IPPO, which primarily optimize individual navigation objectives, lack effective collaborative decision-making capabilities and frequently fail to adjust promptly in potential conflicts, resulting in frequent collisions. While MFPO, CoPO, and TraCo introduce partial interaction modeling, their improvements remain limited due to their primary focus on local (neighborhood) interactions, lacking a global perspective and thus remaining susceptible to systemic congestion or deadlocks.
In the roundabout scenario, COIN also demonstrates strong stability and safety, maintaining smooth trajectories and orderly traffic flow. In contrast, TraCo and CoPO often result in collisions or deviations when vehicles exit the roundabout, while IPPO and ITD3 exhibit poorer stability, leading to frequent collisions during navigation.
In the bottleneck environment, critical interactions mainly occur when vehicles enter the bottleneck region. COIN effectively models both local and global interactions, enabling it to arrange the vehicle passing order appropriately and significantly reducing conflicts and congestion. In the later stages, where interactions become less frequent, COIN maintains highly consistent trajectories with very low error rates, highlighting its capability to balance individual navigation and global cooperation objectives effectively.

\begin{table*}[t!]
\centering
\caption{Results of COIN's ablation study. The best performance is in bold, and the second-best is underlined.}
\resizebox{\textwidth}{!}{
\begin{tabular}{c|cccccc}
\hline
\multirow{2}{*}{Method} & \multicolumn{6}{c}{Intersection environment with 30 initialized agents}                                                                                     \\ \cline{2-7} 
                        & Success Rate $\uparrow$           & Off-Road Rate $\downarrow$             & Collision Rate $\downarrow$           & Efficiency $\uparrow$            & Safety $\uparrow$               & Average Travel Steps $\downarrow$             \\ \hline
COIN w/o GAT            & 81.91 (8.02)           & 6.32 (4.19)          & 11.29 (5.61)         & \textbf{71.04 (19.28)} & -19.25 (7.98)         & \textbf{449.76 (69.02)} \\
COIN w/o VAE            & \underline{84.70 (11.27)}    & \underline{5.06 (4.74)}    & \underline{6.01 (4.75)}    & 64.59 (20.50)          & \underline{-9.45 (6.21)}    & 525.80 (98.12)          \\
COIN w/o VAE+GAT        & 67.56 (10.06)          & 9.74 (5.73)          & 22.74 (9.00)         & 46.41 (25.65)          & -43.25 (13.76)        & 510.41 (75.33)          \\
COIN                   & \textbf{88.78 (10.09)} & \textbf{4.13 (3.90)} & \textbf{3.94 (3.74)} & \underline{68.64 (15.47)}    & \textbf{-6.73 (4.72)} & \underline{509.60 (79.94)}          \\ \hline
                        & \multicolumn{6}{c}{Roundabout environment with 40 initialized  agents}                                                                                       \\ \hline
COIN w/o GAT            & \underline{86.33 (6.25)}     & \underline{6.23 (3.59)}          & 6.06 (3.89)          & \underline{72.92 (12.91)}    & -12.15 (7.98)         & \underline{587.07 (51.97)}    \\
COIN w/o VAE            & 83.51 (6.34)           & 7.13 (4.55)          & \underline{4.36 (4.65)}    & 64.32 (10.00)          & \underline{-10.87 (7.39)}   & 654.33 (49.99)          \\
COIN w/o VAE+GAT        & 80.09 (7.89)           & 10.28 (3.85)         & 8.18 (5.76)          & 64.42 (14.31)          & -20.44 (11.30)        & 612.37 (46.50)          \\
COIN                   & \textbf{90.68 (4.82)}  & \textbf{5.36 (2.99)}    & \textbf{3.22 (3.38)} & \textbf{80.66 (9.46)}  & \textbf{-8.58 (5.03)} & \textbf{579.01 (39.14)} \\ \hline
\end{tabular}
}
\label{table:results_ablation}
\end{table*}

\subsection{Ablation Study}
We conduct an ablation study to assess the effectiveness of each key component of COIN. 
The results shown in Table~\ref{table:results_ablation} demonstrate the importance of both local and global interaction-aware modules. 
With the full interaction-aware design, COIN achieves the highest success rates in both intersection (88.78\%) and roundabout (90.68\%) scenarios. When the global module (COIN w/o GAT) is removed, the success rates drop to 81.91\% and 86.33\%, corresponding to relative declines of 6.87\% and 4.35\%, respectively.
This highlights the critical role of global interactions in agent coordination. 
Removing the local interaction-aware module (COIN w/o VAE) causes a larger decline in success rates, falling to 84.70\% and 83.51\%, with relative decreases of 4.08\% and 7.17\%. 
This underscores the importance of local pairwise interactions in making informed decisions, especially in dense and highly interactive traffic environments.  
The combined removal of both interaction-aware modules (COIN w/o VAE $+$ GAT) results in further performance degradation, with success rates dropping to 67.56\% and 80.09\%, representing relative decreases of 21.22\% and 10.59\%, respectively. 
In summary, these results provide strong empirical evidence that both the local and global interaction-aware modules are critical for improving collaborative learning performance. By explicitly modeling local and global interactions, these modules enable a more expressive centralized critic and facilitate more accurate credit assignment, which in turn lead to more effective agent collaboration and safer, more efficient navigation in dense traffic environments.

\subsection{Real-World Demonstration}

\begin{figure}
    \centering
    \includegraphics[width=\linewidth]{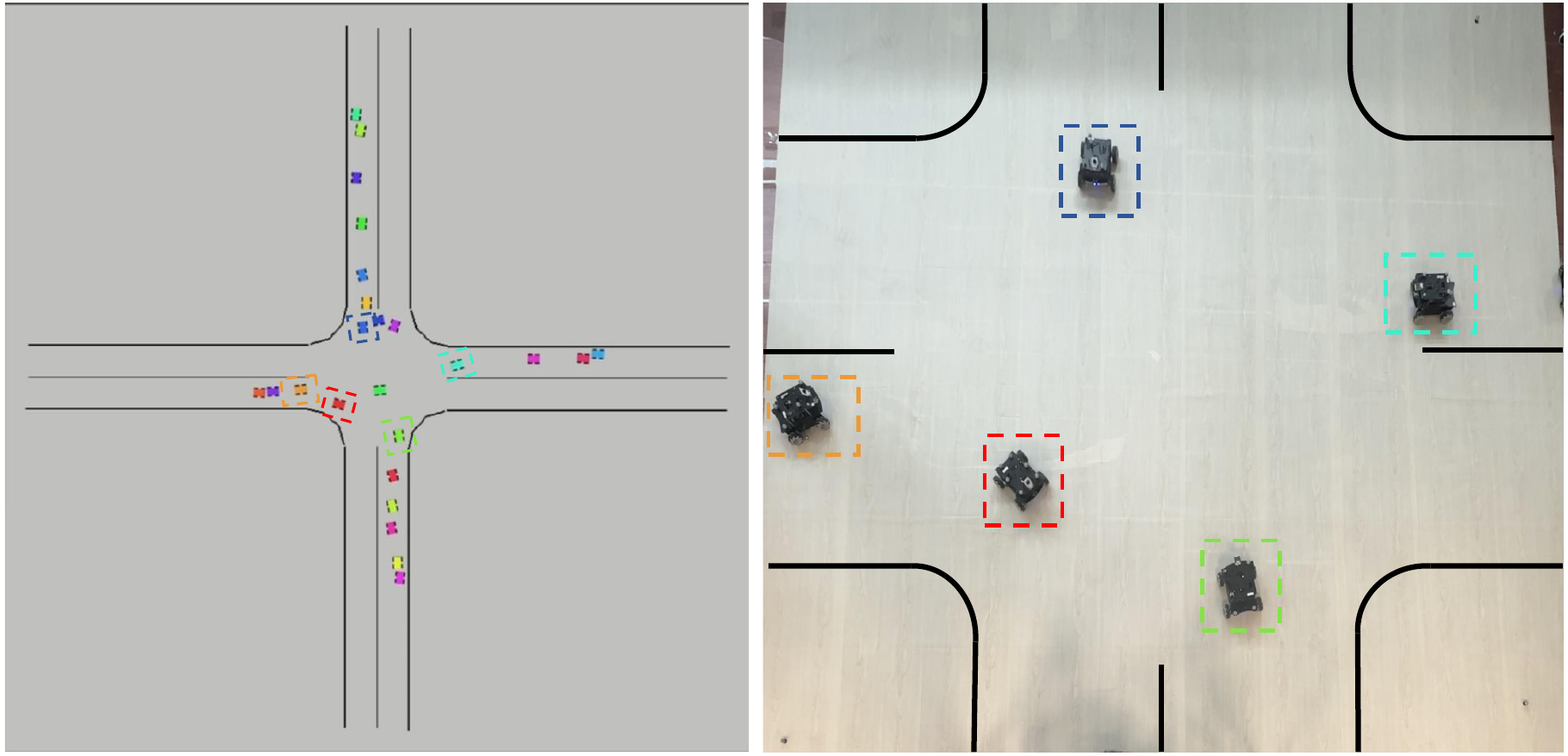}
    \caption{
    Hybrid experiment setup for validating the learned navigation policy. The left side shows a ROS-based simulation where colored blocks represent vehicles. The right side shows a mockup of an unsignalized intersection, where some vehicles are physically represented by mobile robots (highlighted with dashed boxes in matching colors). The real robots follow the trajectories generated by our learned COIN policy.
    }
    \label{fig:real_world_exp}
\end{figure}

To demonstrate the real-world applicability of \textsc{COIN}, we conduct experiments using eight holonomic, mecanum-wheeled robots ($0.2\,\mathrm{m} \times 0.23\,\mathrm{m}$), each equipped with a Jetson Nano 4GB processor, in a mock unsignalized intersection scenario. 
For this demonstration, we adopt the execution framework proposed in~\cite{P3GASUS}. 
We use the \textit{Robot Operating System} (ROS)~\cite{macenski2022robot} for our hybrid simulation, which allows us to send action commands to the robots based on trajectories generated by the trained \textsc{COIN} models, as shown on the left side of Fig.~\ref{fig:real_world_exp}. 
The action commands produced by \textsc{COIN} are used to extract waypoints for all agents. 
These waypoints are then employed to construct a MAGE-like execution graph, as proposed in~\cite{P3GASUS}, which is executed with a planning horizon of three. 
On the right side of Fig.~\ref{fig:real_world_exp}, we present the physical mockup used for real-world validation, where only a subset of vehicle trajectories is executed using real mobile robots. 
These robots are highlighted with dashed boxes in the same colors as their simulated counterparts, while the remaining vehicles exist as virtual agents, however agents cannot differentiate between real and virtual robots. 
For accurate localization during experiments, we employ the \textit{OptiTrack Motion Capture System}, which uses multiple infrared cameras to track reflective markers mounted on each robot, enabling precise estimation of robot positions throughout the experiment. 
The demonstration shows the robots efficiently navigating toward their destinations while cooperating to avoid collisions, highlighting the effectiveness of \textsc{COIN} in complex MASD systems and its potential for end-to-end navigation in realistic and challenging traffic environments.
Full demonstrations, including videos of the real-world multi-robot experiments and comparative results of different algorithms in simulation environments, are available at \url{https://marmotlab.github.io/COIN/}.

\section{Limitations and Future Work}
\label{sec:limitation_future_work}
Although COIN performs well across the evaluated scenarios, several limitations remain and point to valuable directions for future work,  summarized as follows:

Simulation-to-real and large-scale real-world evaluation: 
In this work, our experiments are conducted in the MetaDrive simulator, which provides an interactive, closed-loop environment for studying cooperative multi-vehicle navigation. 
Such closed-loop simulation is essential for RL-based navigation, as it enables continuous feedback and preserves the Markov decision process assumptions required for policy learning and evaluation.
We also include a small-scale hybrid real-robot demonstration to show that the learned policy can be transferred to physical MASD systems.
A natural next step is to carry out full on-board evaluations in real-world multi-vehicle environments, or to design benchmarks that integrate real traffic data with closed-loop decision making.
These extensions would offer a more comprehensive view of COIN’s applicability and robustness in practical deployments.

Low-level control and driving comfort:
COIN adopts an end-to-end control paradigm in which the policy directly outputs low-level control commands, i.e., steering and throttle. 
While this design enables unified learning of perception, planning, and control, it does not impose explicit vehicle-dynamics constraints during training. 
As a result, the learned policy may occasionally generate abrupt or unstable motions and may not always follow lane lines accurately, which can affect driving comfort.
In addition, following common practice in prior work, the reward function used in COIN focuses primarily on safety and efficiency, without explicitly modeling driving comfort or human preferences. 
Developing more adaptive reward formulations that balance these objectives, together with sensitivity analysis of reward weight settings and incorporating dynamics consistency or smoothing mechanisms, is an important direction for improving motion smoothness, comfort, and overall driving quality in future work.

Scalability to larger MASD systems:
This work focuses on dense, highly interactive driving scenarios, where spatial constraints naturally limit the number of agents that can operate simultaneously. Such settings allow us to evaluate cooperative behavior under strong interaction, but they do not cover the broader range of large, city-scale environments. Moving to larger MASD systems brings additional challenges. For example, coordination needs to handle the trade-offs among efficiency, safety and comfort, which become more significant as the network grows. Larger traffic environments also contain more diverse road structures and traffic patterns, which demand stronger generalization and robustness. Extending COIN to these large-scale settings remains an important direction for our future work.

\section{Conclusion}
\label{sec:conclusion}

In this paper, we introduce COIN, a novel collaborative interaction-aware MARL framework designed for end-to-end vehicle navigation in MASD systems.
Specifically, we propose the CIG-TD3 algorithm, which is structured in a CTDE manner to balance individual navigation and global collaboration objectives.
Additionally, we introduce an interaction-aware centralized critic, enhanced by variational inference and graph attention mechanisms, which effectively captures both local and global agent interactions. 
This enables more accurate credit assignment and more effective learning of collaborative strategies, leading to safer and more efficient navigation in dense, highly interactive traffic environments. 
Comprehensive results show that COIN outperforms various baselines in efficiency, safety, and generalizability, highlighting its effectiveness in collaborative navigation and applicability in dynamic MASD systems.
Our real-world robot demonstration further validates COIN’s potential for real-world deployment, showing its effectiveness in dynamic, realistic MASD environments.
Moreover, its flexibility makes it well-suited for other multi-robot navigation systems, particularly in complex, dense environments.

\section*{Acknowledgement}

This work is supported by A*STAR, CISCO Systems (USA) Pte. Ltd and National University of Singapore under its Cisco-NUS Accelerated Digital Economy Corporate Laboratory (Award I21001E0002).

\newpage
\section*{Appendix}
\setcounter{section}{0}
\renewcommand{\thesection}{\arabic{section}}

\setcounter{figure}{0}
\setcounter{table}{0}

\renewcommand{\thefigure}{A.\arabic{figure}}
\renewcommand{\thetable}{A.\arabic{table}}

\section{RL Agent Design}
We define the observation, action, and reward structures for the multi-vehicle collaborative navigation problem, following the settings used in the prior work CoPO~\cite{peng2021learning}.
Further implementation details on the observation, action, and reward design can be found in the official simulator documentation\footnote{\url{https://metadrive-simulator.readthedocs.io/en/latest/}}.

\subsection{Observation}
The observation of each agent consists of two parts: ego states and a 2D lidar point cloud, both measured relative to the ego coordinate frame. 
The ego states include components such as the steering vector, heading vector, the velocity vector, the relative distance to the boundary of roads, and the next two navigation checkpoints of the reference trajectory, totaling 19 dimensions. 
Notably, the reference trajectory is generated at the beginning of each episode and remains fixed for each vehicle, with checkpoints placed at regular intervals along the planned route. 
The two nearest checkpoints are projected into the ego vehicle’s local frame and serve as the important guidance for navigation.
Each agent is also equipped with a lidar to sense its surrounding environment within a 50-meter radius, resulting in a 72-dimensional 2D point cloud vector. 
Thus, the total observation dimension is 91.

\subsection{Action}
The action is defined as a two-dimensional continuous vector $\mathbf{a}=[a_1, a_2] \in [-1, 1]^2$, which is converted into steering (degrees), acceleration (horsepower), and brake signals (horsepower) to control the low-level driving behavior of vehicles. Steering is applied to the two front wheels, while acceleration and brake forces are applied to all four wheels. 
Further implementation details of the action-to-control signal conversion can be found in the simulator documentation\footnote{\url{https://metadrive-simulator.readthedocs.io/en/latest/action.html}}.

\subsection{Reward}
The individual reward function is defined as a combination of two dense (step-wise) driving rewards with a sparse terminal reward, which is formulated as follows:
$r^I=c_1 R_{driving}+ c_2 R_{speed} + R_{termination}$,
where the driving reward $R_{driving}=d_t-d_{t-1}$, with $d_t$ and $d_{t-1}$ denoting the longitudinal coordinates of the vehicle on the current reference lane of two consecutive time steps. 
The speed reward $R_{speed}$ is calculated by normalizing vehicle speed at each time step.
The termination reward $R_{termination}$ includes sparse rewards: $+10$ for successful arrival, $-5$ for running off-road, and $-5$ for collisions. 
The reward components are balanced using the weight factors $c_1=1$ and $c_2=0.1$.
The reward weights $c_1$ and $c_2$ follow the original configuration of the MetaDrive simulator~\cite{li2022metadrive}, which has been widely adopted in prior work and is empirically shown to provide a reasonable and effective learning signal for multi-vehicle cooperative navigation tasks.
In addition, the global reward $r^G$ is defined as the average of individual rewards across all active agents in the system at each timestep, serving as a system-level signal to encourage overall coordination and collective performance.
This design helps balance individual navigation objectives with collaborative behavior, thereby improving both operational efficiency and agent interaction, especially in dense or conflict-prone environments.

\section{Experimental Results}

\subsection{Additional Results of Local Interaction-Aware Modules}

The local interaction-aware module in the centralized critic is designed to capture how pairwise agent interactions affect the ego agent’s future observation-state transitions and rewards. 
In dense and highly-interactive MASD scenarios, such effects are primarily reflected in future system evolution rather than instantaneous feature correlations. 
To address this, we adopt a CVAE trained to reconstruct the ego agent’s future observation, state, and reward conditioned on both agents’ current observations, states, and actions. 
Thus, the resulting latent variable $z_{ij}$ encodes interaction-related variations in future transitions, providing a predictive interaction representation rather than a deterministic fusion of current features.

To empirically validate this design choice, we conduct an additional comparison experiment in which the CVAE-based local interaction-aware module is replaced with an attention-based alternative using a graph attention mechanism for pairwise feature fusion. 
All other components and training settings are kept identical. 
Both variants are trained with eight random seeds and evaluated over 160 episodes in total. 
As shown in Table~\ref{table:results_local_interaction_module}, the CVAE-based model consistently outperforms the attention-based variant across all evaluation metrics, indicating that explicitly modeling interaction-specific future effects yields enhanced performance and more effective centralized value estimation in dense MASD scenarios.

\begin{table*}[t!]
\centering
\caption{
Comparison of COIN with the original CVAE-based local interaction-aware module and an attention-based variant in the intersection environment. 
The best results are shown in bold.
}
\resizebox{\textwidth}{!}{
\begin{tabular}{c|ccccc}
\hline
\multirow{2}{*}{Method} & \multicolumn{5}{c}{Intersection environment with 30 initialized agents}                                                                                     \\ \cline{2-6} 
                        & Success Rate $\uparrow$           & Off-Road Rate $\downarrow$             & Collision Rate $\downarrow$           & Efficiency $\uparrow$            & Safety $\uparrow$                      \\ \hline
COIN (original) & \textbf{88.78 (10.09)} & \textbf{4.13 (3.90)} & \textbf{3.94 (3.74)} & \textbf{68.64 (15.47)}    & \textbf{-6.73 (4.72)}  \\ \hline
COIN (attention) & 85.92(11.26) & 5.88 (5.67)  & 6.34 (6.50) & 57.35 (23.85) &  -9.95 (10.10)  \\ \hline
\end{tabular}
}
\label{table:results_local_interaction_module}
\end{table*}

We further analyze the learned latent representations produced by the local interaction-aware module.
Specifically, we run multiple evaluation episodes using the trained COIN model in each of the three environments and collect pairwise latent variables $z_{ij}$ across different agents and time steps, resulting in totaling 1500 latent samples per environment.
We then apply clustering to the collected latent variables and visualize their distribution using t-SNE~\cite{maaten2008visualizing}.

\begin{figure*}[t!]
    \centering
    \includegraphics[width=\linewidth]{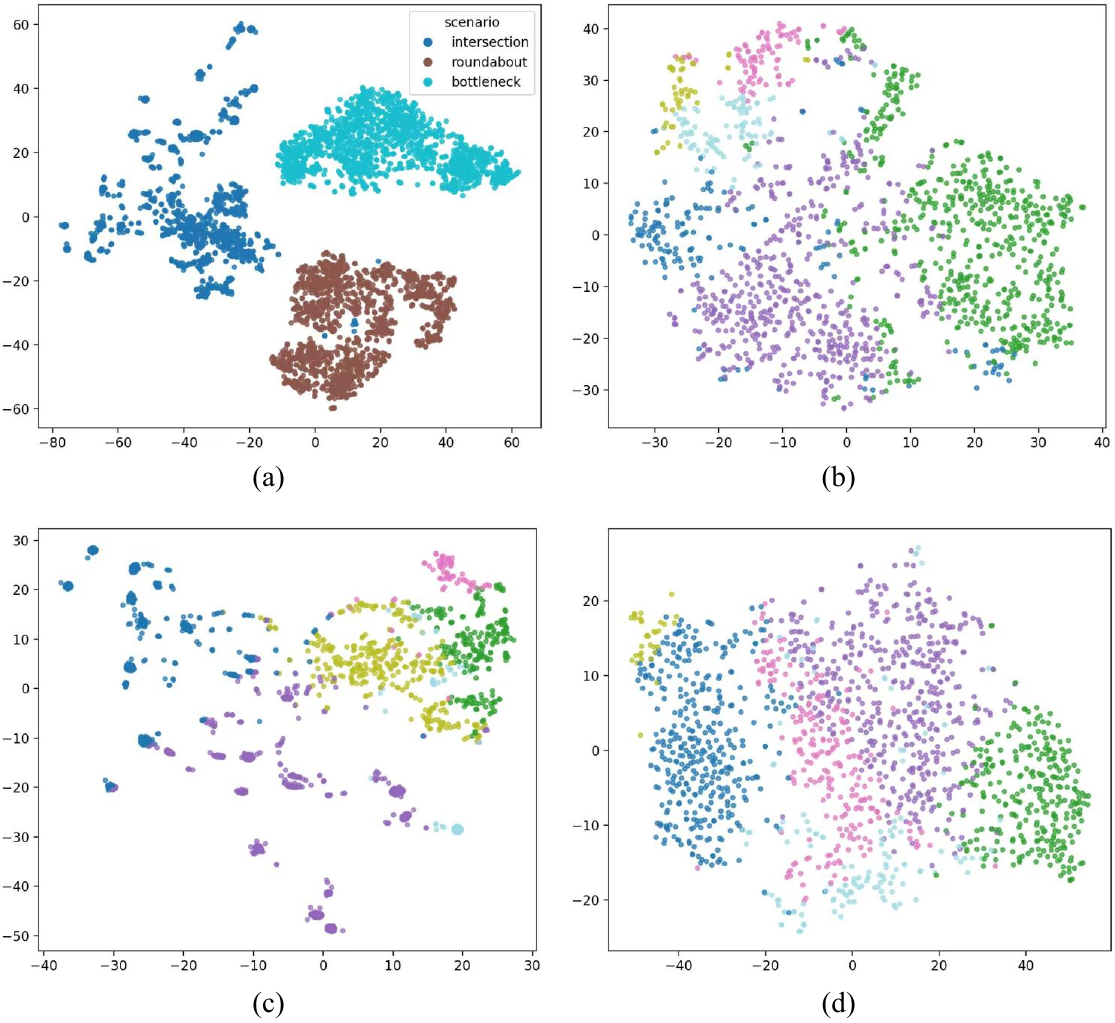}
    \caption{
    t-SNE visualization of the learned latent variable $z_{ij}$. (a) Latent representations learned by COIN’s local interaction-aware module across three traffic environments. (b)–(d) Latent representations in the roundabout, intersection, and bottleneck scenarios, respectively, where the samples are grouped into six clusters.
    }
    \label{fig:tsne_results}
\end{figure*}

Fig.~\ref{fig:tsne_results}(a) shows the t-SNE visualization of latent variables aggregated across all three environments.
We observe that the latent representations from different scenarios are clearly separated and form three distinct groups, indicating that interaction patterns vary across scenarios and are effectively captured by the proposed CVAE-based local interaction-aware module.
Figs.~\ref{fig:tsne_results}(b)–(d) present t-SNE visualizations for the roundabout, intersection, and bottleneck scenarios, respectively, where latent variables are clustered in to six categories prior to projection.
Although we do not explicitly enforce an alignment between these latent variables and semantic behavior categories, the visualizations exhibit clear and well-separated clusters, suggesting that the CVAE adaptively organizes interaction-dependent patterns in the latent space.
Overall, the results indicate that the learned latent representations are diverse and structured, rather than collapsed, and encode interaction-specific variations across different driving scenarios.

\subsection{Additional Results of Global Interaction-Aware Modules}

Our work focuses on dense, spatially constrained traffic scenarios, such as intersections, where coordination often depends on system-level dependencies that are difficult to capture using strictly local neighborhoods. 
To model such broader interactions, we adopt a fully connected graph in the global interaction-aware module of the centralized critic under the CTDE paradigm. 
This design affects the training phase only. 
During execution, each agent acts solely based on local observations without constructing a global graph or requiring additional communication, so deployment efficiency and runtime scalability are not impacted.
In these settings, agents that are not immediate neighbors, for example vehicles approaching a conflict region, can influence current decisions through queue propagation and multi-step interaction effects, which motivates the use of a fully connected graph in the centralized critic.

The additional training cost introduced by the full graph remains manageable at the considered scale, as the number of simultaneously interacting vehicles is naturally limited by road geometry.
Moreover, we adopt the respawn mechanism in~\cite{peng2021learning} to maintain persistent high-density traffic over time rather than assuming a fixed agent population within an episode. 
To further examine this design choice, we compare the fully connected graph with a radius-based alternative (e.g., 40m) using the same GAT architecture. 
As shown in Table~\ref{table:results_global_interaction_module_graph}, the fully connected graph consistently outperforms the radius-based variant across all metrics, indicating that modeling broader-range interactions leads to improved performance and more accurate and globally consistent value estimation in dense intersection scenarios.

\begin{table*}[t!]
\centering
\caption{
Comparison of COIN with full-graph and radius-based (40 meters) sparse graph global interaction-aware modules under the intersection environment. The best results are shown in bold.
}
\resizebox{\textwidth}{!}{
\begin{tabular}{c|ccccc}
\hline
\multirow{2}{*}{Method} & \multicolumn{5}{c}{Intersection environment with 30 initialized agents}                                                                                     \\ \cline{2-6} 
                        & Success Rate $\uparrow$           & Off-Road Rate $\downarrow$             & Collision Rate $\downarrow$           & Efficiency $\uparrow$            & Safety $\uparrow$                      \\ \hline
COIN (Full-Graph) & \textbf{88.78 (10.09)} & \textbf{4.13 (3.90)} & \textbf{3.94 (3.74)} & \textbf{68.64 (15.47)}    & \textbf{-6.73 (4.72)}  \\ \hline
COIN (Sparse-Graph) & 83.30 (11.73) & 5.34 (3.85)  & 5.95 (4.96) &  58.60 (18.43) &  -8.83 (4.44)  \\ \hline
\end{tabular}
}
\label{table:results_global_interaction_module_graph}
\end{table*}

In dense traffic scenarios, the influence of neighboring agents on global value estimation is non-uniform and varies over time.
Uniform aggregation schemes such as GCN~\cite{kipf2016semi} and GraphSAGE~\cite{hamilton2017inductive} implicitly assume similar contributions from all neighbors and may weaken the impact of critical interactions. 
In contrast, GAT assigns adaptive weights to neighbors, enabling the centralized critic to focus on coordination- and safety-critical interactions among agents.

We further conduct an ablation study by replacing the GAT in the global interaction-aware module with GCN and GraphSAGE while keeping all other components unchanged.
As shown in Table~\ref{table:results_global_interaction_module_gat}, the GAT-based design consistently achieves better performance across key evaluation metrics, including success rate, collision rate, and safety.
While GCN and GraphSAGE reduce model complexity, they lead to degraded global interaction modeling in the intersection environment, with GraphSAGE in particular exhibiting higher collision rates despite improvements in some efficiency-related metrics. 
This suggests that GAT is more suitable for our global interaction-aware modeling, as its attention mechanism allows the centralized critic to emphasize critical interactions in the complex and dynamic MASD systems.

\begin{table*}[t!]
\centering
\caption{
Comparison of COIN with different graph encoders (GAT, GCN, and GraphSAGE) used in the global interaction-aware module under the intersection environment. The best results are shown in bold.
}
\resizebox{\textwidth}{!}{
\begin{tabular}{c|ccccc}
\hline
\multirow{2}{*}{Method} & \multicolumn{5}{c}{Intersection environment with 30 initialized agents} \\ \cline{2-6}
 & Success Rate $\uparrow$ & Off-Road Rate $\downarrow$ & Collision Rate $\downarrow$ & Efficiency $\uparrow$ & Safety $\uparrow$ \\ \hline
COIN (GAT) 
& \textbf{88.78 (10.09)} 
& 4.13 (3.90) 
& \textbf{3.94 (3.74)} 
& 68.64 (15.47) 
& \textbf{-6.73 (4.72)} \\ \hline
COIN (GCN) 
& 83.32 (12.75) 
& 7.31 (6.20) 
& 4.74 (4.70) 
& 57.56 (17.85) 
& -10.06 (7.32) \\ \hline
COIN (GraphSAGE) 
& 84.53 (8.56) 
& \textbf{3.57 (2.67)} 
& 10.61 (6.02) 
& \textbf{78.64 (22.52)} 
& -15.36 (7.74) \\ \hline
\end{tabular}
}
\label{table:results_global_interaction_module_gat}
\end{table*}

\newpage
\section*{References}
\bibliographystyle{elsarticle-harv}
\bibliography{ref}

\end{document}